\documentclass[acmlarge]{acmart}
 \makeatletter
\newcommand{\myconfshort}{\acmConference@shortname}
\newcommand{\myconffull}{\acmConference@name}
\newcommand{\myconfdate}{\acmConference@date}
\newcommand{\myconfloc}{\acmConference@venue}
\AtBeginDocument{
  \fancypagestyle{firstpagestyle}{
    \fancyhead{}%
    \fancyfoot[C]{}%
  }
  \fancyhf{}
  \fancyhead[LO]{\@headfootfont\shorttitle}%
  \fancyhead[RE]{\@headfootfont\@shortauthors}%
  \fancyhead[LE]{\@headfootfont\footnotesize \myconfshort, \myconfdate, \myconfloc}%
  \fancyhead[RO]{\@headfootfont\footnotesize \myconfshort, \myconfdate, \myconfloc}%
  \fancyfoot[C]{}%
}
\makeatother
\acmBooktitle{\conffull\@ (\confshort), \confdate, \confloc}
\AtBeginDocument{%
  }

\setcopyright{acmlicensed}
\copyrightyear{2026}
\acmYear{2026}
\setcopyright{cc}
\setcctype{by}
\acmConference[FAccT '26]{The 2026 ACM Conference on Fairness, Accountability, and Transparency}{June 25--28, 2026}{Montreal, QC, Canada}
\acmBooktitle{The 2026 ACM Conference on Fairness, Accountability, and Transparency (FAccT '26), June 25--28, 2026, Montreal, QC, Canada}
\acmDOI{10.1145/3805689.3812333}
\acmISBN{979-8-4007-2596-8/2026/06}

\usepackage{wrapfig}
\usepackage{subcaption}
\usepackage{multirow}
\usepackage{pifont}
\usepackage{tabularray}
\newcommand{\cmark}{\ding{51}}

\usepackage{microtype}
\usepackage{hyperref}
\usepackage{url}
\usepackage{booktabs}
\usepackage{graphicx}
\usepackage{CJKutf8}
\usepackage{listings}
\usepackage{lineno}
\usepackage{multirow}
\usepackage{xcolor}
\definecolor{darkblue}{rgb}{0, 0, 0.5}
\hypersetup{colorlinks=true, citecolor=darkblue, linkcolor=darkblue, urlcolor=darkblue}
\begin{document}

\title{Do Chinese models speak Chinese languages?}

\author{Andrea Wen-Yi Wang}
\authornote{Equal contribution}
\email{aww66@cornell.edu}
\author{Unso Eun Seo Jo}
\authornotemark[1]
\email{unsojo@cornell.edu}
\affiliation{%
  \institution{Cornell University}
  \city{Ithaca}
  \state{NY}
  \country{USA}
}

\author{David Mimno}
\affiliation{%
  \institution{Cornell University}
  \city{Ithaca}
  \country{USA}}
\email{mimno@cornell.edu}

\received{13 Jan 2026}
\received[revised]{25 March 2026}
\received[accepted]{16 April 2026}

\begin{abstract}
The release of top-performing open-weight LLMs has cemented China's role as a leading force in AI development.
Do these models support languages spoken in China? Or do they support the same languages as models developed in the United States or in Europe? 
Comparing multilingual capabilities is important for two reasons.
First, language ability provides insights into pre-training data curation, and thus into resource allocation and development priorities.
Second, Chinese model developers need to navigate the tension between serving a linguistically diverse population domestically, and optimizing for globally visible benchmarks that are predominantly English. 
We investigate Chinese model developers' priorities through a comparative study of Chinese-developed and Western-developed open-weight LLMs, on 21 language variants including Asian regional, Chinese, and European languages. Our experiments on Information Parity and reading comprehension show Chinese models' performance across these languages correlates strongly (r=0.93) with their Western counterparts, with the sole exception being better Mandarin. Chinese-developed models are good at French and German, but they sometimes cannot identify languages spoken by Chinese minorities such as Kazakh and Uyghur. 
Overall, all open-weight LLMs we study have a similar multilingual performance profile, despite the diverse linguistic and cultural contexts the model developers operated within. We interpret the homogenization as consistent with the influence of global benchmarking practices and shared training resources. Rather than treating current language support as inevitable, our results highlight multilingual development as a space of prioritization and trade-offs, with implications for model developers, policymakers, and users.
\end{abstract}

\maketitle
\section{Introduction}
Large language models (LLMs) are prohibitively expensive to develop, pushing model developers to make decisions on resource allocations. 
For developers located in complex linguistic environments, this means choosing to prioritize some languages over others. 
In this paper, we study a salient example: China.
Domestically, China is a linguistically diverse country with 1.4 billion people speaking Mandarin and hundreds of other languages and dialects, but not English \citep{ethnologue-cn, Erard2009HowMany}.
At the same time, for Chinese-developed LLMs to gain global attention, developers are incentivized to optimize for popular evaluation benchmarks that are predominantly English \citep{deepseek-r1}. 
Therefore, China's model developers need to navigate the tension between serving a linguistically diverse local population or optimizing for performance on globally visible benchmarks in English. 

In this paper, we first propose four ways in which Chinese developers could have chosen to prioritize linguistic diversity as informed by China's history, policy, and geopolitical context.
Then, we conduct computational experiments to probe which prioritization options China's model developers seem to have adopted. Specifically, we compare 6 Chinese- and 4 Western-developed open-weight LLMs's multilingual performances on 21 language variants, including Mandarin Chinese, Chinese minority languages, Asian regional languages, and European languages. We evaluate multilingual performance using both task-agnostic and task-dependent experiments, measuring Information Parity, machine reading comprehension, and language identification. 

Given data disparity, we do not expect LLMs to do as well in minority languages, such as Kazakh, as in Mandarin. The question we ask is \textit{not} whether Chinese-developed LLMs (such as \texttt{Qwen}) perform similarly across languages. Instead, we ask whether Chinese-developed LLMs are \textit{comparatively better than} their Western counterparts (such as \texttt{Llama}) on Mandarin and minority languages such as Kazakh, given the assumption that China's model developers would have more access to these domestic languages than their Western counterparts. 

Overall, we find that multilingual performance across the 21 language variants we examine is highly correlated between average Chinese-developed and average Western-developed open-weight LLMs. One notable exception is Mandarin, in which Chinese-developed models perform better than their Western counterparts. Beyond this difference, all popular open models we look at, regardless of origins, correlates highly with each other in terms of multilingual performance. 

Our findings suggest that China's model developers are optimizing for popular evaluative benchmarks to gain global visibility on their models. They appear to be making similar decisions on allocating model capacity and resources across languages, resulting in highly similar multilingual performance profiles with their Western counterparts.  Additionally, our findings reveal a pattern of homogenization in multilingual performance across the models we study, even when model developers operate in distinct linguistic, political, and cultural contexts. This suggests that there are certain normalizing pressures applied on LLM training via global benchmarking practices and shared resources such as training and evaluation data. 

\begin{table*}
\centering
\small
\begin{tabular}{|l|l|p{0.12\linewidth}|p{0.12\linewidth}|p{0.12\linewidth}|}
\hline
Category & Language & Experiment 1: FLORES+ & Experiment 2: Belebele & Experiment 3: MC$^2$\\
\hline
\hline
\multirow{2}{*}{Mandarin Chinese}&Mandarin (Simplified) & \cmark & \cmark & -\\
& Mandarin (Traditional) & \cmark & \cmark & -\\
\hline
Chinese Han Dialects (Other)&Yue (Cantonese)&\cmark&-&-\\
\hline
\multirow{5}{*}{Chinese Minority}&Jingpho&\cmark&\cmark&-\\
&Lhasa Tibetan&\cmark&\cmark&\cmark\\
&Uyghur&\cmark&-&\cmark\\
&Mongolian&-&-&\cmark\\
&Kazakh&-&-&\cmark\\
\hline
\multirow{2}{*}{Northeast Asian}&Korean&\cmark&\cmark&-\\
&Japanese&\cmark&\cmark&-\\
\hline
\multirow{6}{*}{Southeast Asian}&Indonesian&\cmark&\cmark&-\\
&Lao&\cmark&\cmark&-\\
&Burmese&\cmark&\cmark&-\\
&Thai&\cmark&\cmark&-\\
&Vietnamese&\cmark&\cmark&-\\
&Standard Malay&\cmark&\cmark&-\\
\hline

\multirow{5}{*}{European}&English&$\star$&\cmark&-\\
&French&\cmark&\cmark&-\\
&Italian&\cmark&\cmark&-\\
&Spanish&\cmark&\cmark&-\\
&German&\cmark&\cmark&-\\
\hline
\end{tabular}
\caption{Languages evaluated in each experiment. In Experiment 1, we evaluate Information Parity (IP) on FLORES+ Machine Translation benchmark dataset \citep{goyal-etal-2022-flores, nllb-24}. In Experiment 2, we evaluate Machine Reading Comprehension (MRC) on the Belebele Dataset \citep{bandarkar-etal-2024-belebele}. In Experiment 3, we utilize Multilingual Corpus of Minority Languages in China (MC$^2$) dataset to evaluate if models' could identify Chinese minority languages included in the dataset \citep{zhang-etal-2024-mc2}. We categorize languages used by populations in China that the PRC determines as ``minority ethnic group'' as ``Chinese Minority''. We evaluate the Kazakh in Kazakh Arabic script and Mongolian in Traditional Mongolian script, which are writing systems used in China. $\star$: English is used as the reference language \citep{tsvetkov-kipnis-2024-information}.} 
\label{tab:langs}
\end{table*}

\section{Research Background: Language and AI Policy in China}
\paragraph{Language Policy in China} 
For thousands of years, central governments in
China have used language as a tool to manage a
vast multiethnic population that speaks over a hundred languages and dialects. These policies
have alternated from being pluralist to assimilationist
over the eras. One of the earliest such mandates, in 221 B.C., was the First Emperor of Qin's program to standard the Chinese script as a move to consolidate central state power after unifying the warring states. More recently, following the 1949 Communist Revolution, the Chinese Communist Party (CCP) launched extensive linguistic campaigns to unite the minority population of over 106 million people who spoke 129 languages among them to build an inclusive Chinese nation \citep{mullaney2011coming}.\footnote{Also known as ``Multinational state building''}

China adopted a more assimilationist approach in the late twentieth century with a monolingual policy (``one nation, one language'') that promoted Mandarin as the ``super language'' in a 1982 constitutional amendment.\footnote{In 1982, constitutional amendment Article 19 made Mandarin the official common spoken ``super language'' and in 2000, the People Republic of China (PRC) promulgated the Standard Spoken and Written Chinese Language law to promote Mandarin use in public.}
Yet, digital frontiers have seen grassroots efforts to promote linguistic inclusivity. For example, the 2005 convening of the \textit{National Conference on the Standardization and Computerization of Minority Languages and Writing}, and publications such as the ``Ethnic Language Edition of the Linux Operating System and Office Suite'' and ``Advances in China’s Minority Language Processing'' signal these attempts.

\paragraph{AI Policy in China} 
The PRC government has been taking steps to regulate LLMs and generative AI. In July 2023, China's Cyberspace Administration issued the Administration of Generative AI Services (the ``Interim AI Measures''), requiring generative AI services with ``public opinion attributes'' or ``social mobilization capabilities'' undergo rigorous security assessments. 

The Interim AI Measures require that all AI-generated content comply with five principles such as upholding socialist values, preventing discriminatory content, and implementing transparency and reliability measures. 
As of January 2025, 302 generative AI services had completed the mandatory government filing process to comply with these requirements \citep{cac-filing}.
Indeed, users have reported that models like \texttt{Deepseek-R1} refused to engage with certain topics deemed sensitive by the Chinese government such as Taiwan, Tibet, and Tiananmen Square and instead digress to chat about math, coding, and logic \citep{wired-deepseek-censorship}. 

Data is central to these regulation efforts. When training AI models, the Interim AI Measures require providers to use lawfully sourced data and avoid infringing on intellectual property rights. They must also employ measures to enhance training data quality, truthfulness, accuracy, objectivity, and diversity and comply with national laws.\footnote{Chinese national Cybersecurity Law, Data Security Law, and Personal Information Protection} These requirements are broadly defined and can include minority languages, but there is no explicit mention of language and cultural inclusivity. In other words, there is no evidence of top-down pressure for Chinese organizations to commit resources for minority languages. In 2017, China announced the New Generation AI Development Plan to lead China to become a `major AI innovation center’ by 2030. In this  `New Generation’ AI era, what is China’s AI language policy and how does it relate to minority languages?
\section{Related Work} \label{sec:related-work}
\paragraph{Multilingual LLMs and Sovereign AI}
LLMs have grown increasingly multilingual over the last few years \citep{conneau-etal-2020-unsupervised,brown2020language,workshop2022bloom}. As multilingual LLMs have integrated into the fabric of society, recent scholarship has argued that multilingual LLMs should be understood not merely as technical artifacts but as social actors embedded in broader linguistic, cultural, and institutional contexts \cite{agarwal2025-FluentForeignEven}. Scholars have discussed both the benefits and harms of multilingual LLMs. 
On the one hand, multilingual language technologies are used to improving access to information\citep{lee2020EFL} and preserving low-resource languages \citep{bird-chiang-2012-machine}. Community-led initiatives such as New Zealand’s Te Hiku Media and Africa's Masakhane illustrate the use of language technology to preserve languages not under the spotlight of global LLM development \citep{wired-maori,nekoto-etal-2020-participatory, adelani-etal-2023-masakhanews}. On the other hand, scholars in AI ethics and fairness have also examined the \textit{harm} of these models, particularly concerning their disparate performance among diverse languages and dialects \cite{bella2024-TacklingLanguageModellinga}, and the cultural bias they exhibit \cite{tao2024-CulturalBiasCultural}. These influences are especially consequential as LLMs are used to mediate access to service and make high-stakes decisions such as hiring \cite{Harvey2025-AFramework, Lyu2025CharacterizingBias}. Recognizing the ramifications of LLMs, governments and language communities have invested in the development of sovereign and regional AI: LLMs designed to support local languages and align with local cultural and social values \cite{Alduhishy2024-SovereignAIWhat}. Studies have found that while these LLMs tend to perform better in local languages, they continue to struggle with cultural alignment \cite{agarwal2025-FluentForeignEven, chae2025-AssessingSocioCulturalAlignment}, suggesting that linguistic performance is a more tractable objective compared with cultural alignment. 

\paragraph{Benchmarking}
Benchmark datasets drive modern AI development. While the original purpose of benchmarking practice was to serve as a common standard to objectively evaluate machine learning development, scholars have raised concerns regarding their purported validity. For example, \citet{orr2024-AISportCompetitive} observed that popular benchmark datasets were constructed through arbitrary processes. They claim benchmark datasets for AI development become competitive sports-like platforms but ``fall short in encapsulating the depth and breadth of the tasks they are intended to measure''. \citet{koch2021-ReducedReusedRecycled} observed that developers of popular benchmarks were highly constrained to a select few institutions, raising questions about whether AI systems optimizing for these benchmarks would meaningfully support a more diverse group of users. Albeit these criticisms on benchmarking practices, benchmarks have gained increasing attention outside of the academic community, as both CEOs of Google and Meta boasted their models' performance on popular benchmarks such as MMLU as a token of capability \cite{keegan2024-EveryoneJudgingAI}.
Therefore, model developers have strong incentives to optimize for these benchmarks to gain public recognition for their models.

In this context, we bridge these benchmark pressures with practical local language support pressures in China's LLM development. Specifically, we ask whether China's LLM development priorities align more closely with that of ``sovereign AI'' or optimize for attention on global benchmarks. 

\begin{table*}[t]
\begin{tblr}
{
  cells={valign=m,halign=c},
  row{1}={font=\bfseries,rowsep=8pt},
  colspec={|p{0.13\linewidth}|p{0.12\linewidth}|p{2cm}|p{4cm}|p{3cm}|},
  hlines,
  vlines
}
\centering
\small
Model & Vocabulary Size& Pretrained Data Size (trillion tokens) & Pretrained Data Source & Pretrained Data Languages\\
\hline

\texttt{Qwen2.5}  &  151,643 & 18 & ``Our dataset is designed to meet these requirements and includes public
web documents, encyclopedia, books, codes, etc. Additionally, our dataset is multilingual, with a
significant portion of the data being in English and Chinese.''  (Qwen2) & Multilingual data, with a significant portion in English and Chinese. (Qwen2)\\
\texttt{Yi-1.5} & 64,000 & 3.1 & Common Crawl \footnote{https://commoncrawl.org/} (80\%), encyclopedia, books, papers, codes. (Yi) & English and Chinese (Yi) \\
\texttt{DeepSeek-R1} & 129,280 & 14.8 (DeepSeek-V3) & 
\textit{Unclear}
& ``multilingual coverage beyond English and Chinese" (DeepSeek-V3)\\
\texttt{InternLM3} & around 92,000 
(InternLM2) & 4 & ``The text data in our pre-training dataset can be categorized by source into web pages,
papers, patents, and books. [...] Our web page data mainly comes from Common Crawl.'' (InternLM2) & ``The Chinese and English data from web pages account
for 86.46\% of the total, making it the primary source. '' (InternLM2) \\
\texttt{Baichuan2} & 125,696 & 2.6 & Web pages, books, research papers, codebases, etc. & \textit{Unclear} \\
\end{tblr}
\caption{Training data details according to technical reports and model cards of the models. Many models are updates of earlier pretrained models and lack clear details on pretraining data. In such cases, we refer to the technical reports and model cards of the earlier models. Notably, \texttt{DeepSeek} models have been extremely terse about their pretraining data recipe for both the R1 and V3.}
\label{tab:models}
\end{table*}

\section{Priority Options}
Informed by China's historical use of language as a political tool and recent developments in its AI policy, we identify four priority options. In Table~\ref{tab:models} we identify information included in the technical reports of Chinese open-weight LLMs. In the reports, there is limited information on which languages were used in the pretraining data. Although details remain unclear, most models tend to be explicitly focused on English and Chinese data found through online sources such as Common Crawl and other web pages. Given this limited information, we propose four hypotheses:

\begin{itemize}
    \item \textbf{Null Hypothesis --- }Chinese model developers prioritize optimizing for popular, globally visible benchmarks. Therefore, they are making decisions in data collection and use of datasets similar as model developers elsewhere. Hence, there is no difference in multilingual support between Chinese and Western models.
    \item \textbf{Mandarin Hypothesis ---} China's model developers are incentivized to support local audience, and Mandarin Chinese is the most efficient way to reach a broad domestic audience. Chinese organizations allocated additional resources to improve Mandarin performance, incentivized by socio-political context and ease of access to Mandarin data. In this context, we expect to see Chinese LLMs better than Western ones at Mandarin but not at other local languages. 
    
    \item \textbf{Pluralist Hypothesis --- }Chinese model developers are incentivized to support the local audience through their local languages. While Mandarin is the dominant language, recently more Chinese companies are responding to the growing popularity for dialects and non-Mandarin languages on social media and technology platforms \citep{SocialMediaChineseDialects,li2024chineseDialect,blossomDialect}.\footnote{There is increasing demand for machine translation between Mandarin Chinese and China's minority languages, such as Tibetan, Mongolian, and Uyghurs, to foster the ``the political, economic, and cultural exchanges'' between China and their minority populations \citep{Zhang2024NeuralMachineTranslation}.}
    As China's developers also have more access to a range of local language data compared to Western developers, they are making decisions about data that emphasize more local languages.  Under this hypothesis, we expect to see that Chinese-developed models are better than Western models at Mandarin and other local languages spoken in China.

    \item \textbf{Regional Hypothesis ---} Chinese model developers are incentivized to prioritize languages of larger populations and market power in greater Asia, such as Korean and Japanese. Historically, China and the greater Asian region share linguistic and cultural commonalities. China has historically been a regional ``Middle Kingdom" with influence in the greater Asian regional ``tributary states" such as Korea and Vietnam. Also, many modern scientific and social vocabulary are shared pan-Asia via translation of western concepts. For instance, many Asian languages adopted Japanese western scientific terms (``wasei-kango'': Japanese-made Chinese words) such as physics (butsurin/\begin{CJK*}{UTF8}{mj}物理\end{CJK*}) or phone (denwa/\begin{CJK*}{UTF8}{mj}電話\end{CJK*}). 
    In addition to shared linguistic commonalities, we hypothesize that Chinese model developers have economic incentive to address languages of larger populations or market power in greater Asia, such as Korean and Japanese.
    Under this hypothesis, we expect to see Chinese models are better than Western models in Mandarin and other languages spoken in the greater East and Southeast Asian regions but not in local Chinese minority languages. 
\end{itemize}

\begin{figure*}
\begin{subfigure}{0.48\textwidth} 
\includegraphics[width=\textwidth]{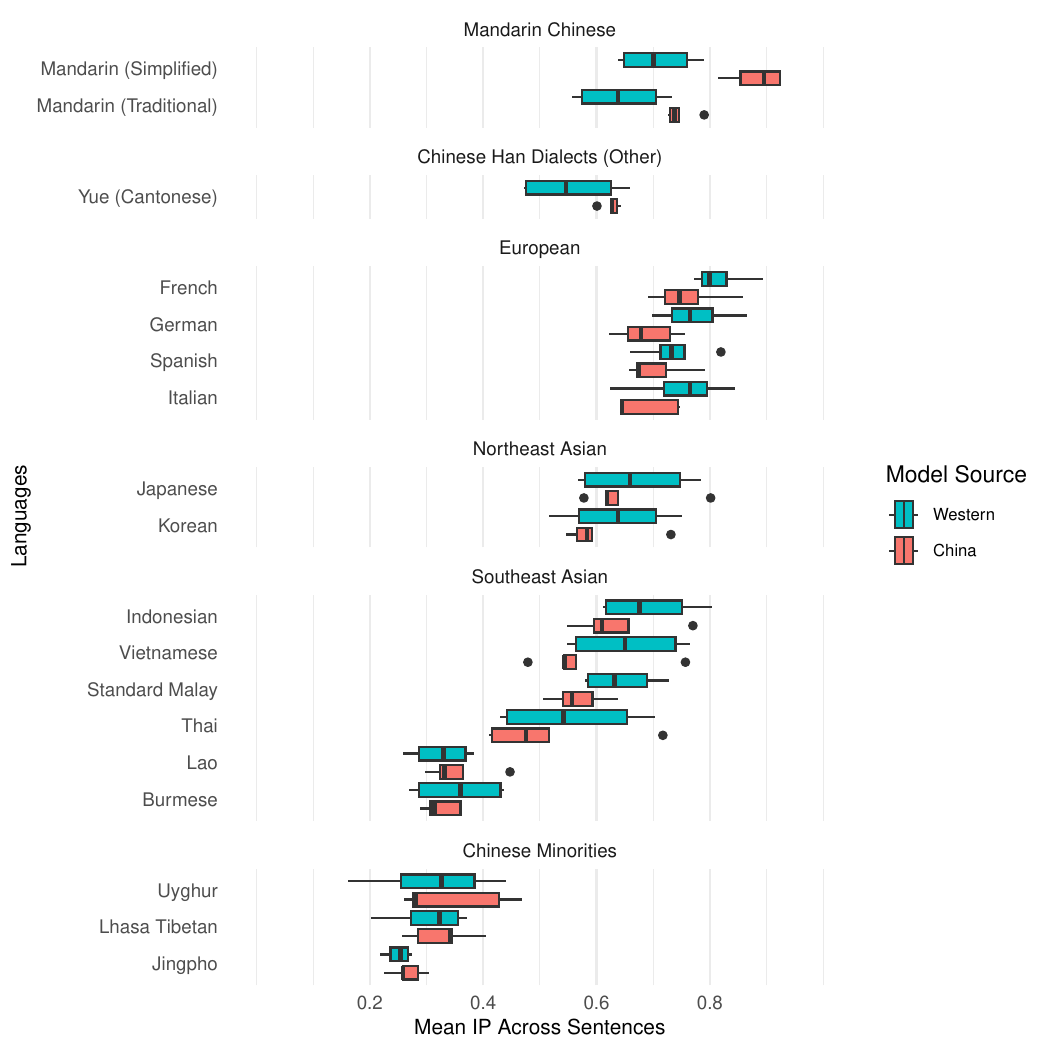}
\caption{Instruction-tuned models}
\label{fig:floresp_ip_whisker_full_chat}
\end{subfigure}
\hfill
\begin{subfigure}{0.48\textwidth} 
\includegraphics[width=\textwidth]{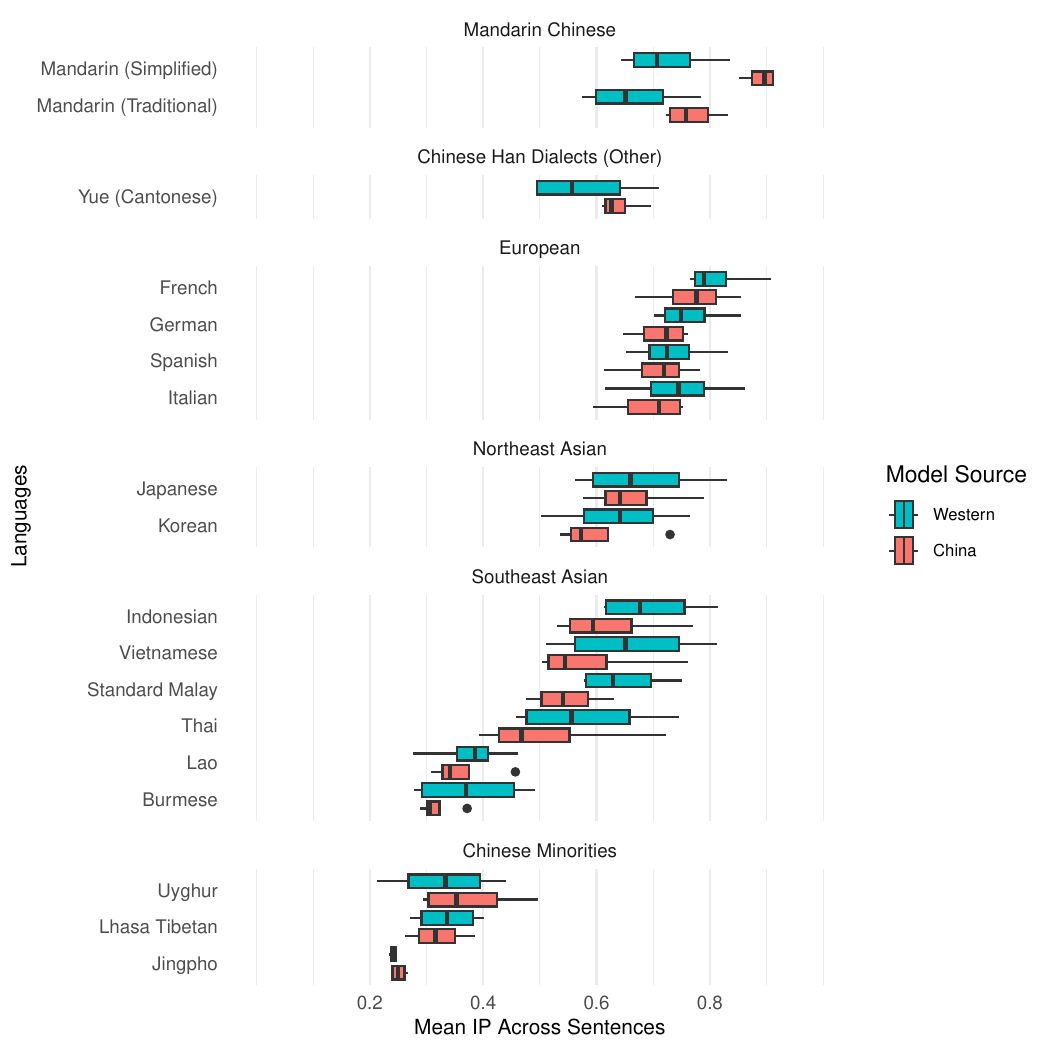}
\caption{Base models}
\label{fig:floresp_ip_whisker_full_base}
\end{subfigure}
\caption{Information Parity (IP) of Chinese-developed v.s. Western-developed models.}
\label{fig:floresp_ip_whisker_full}
\end{figure*}

\begin{figure*}
\begin{subfigure}{0.48\textwidth} 
\includegraphics[width=\textwidth]{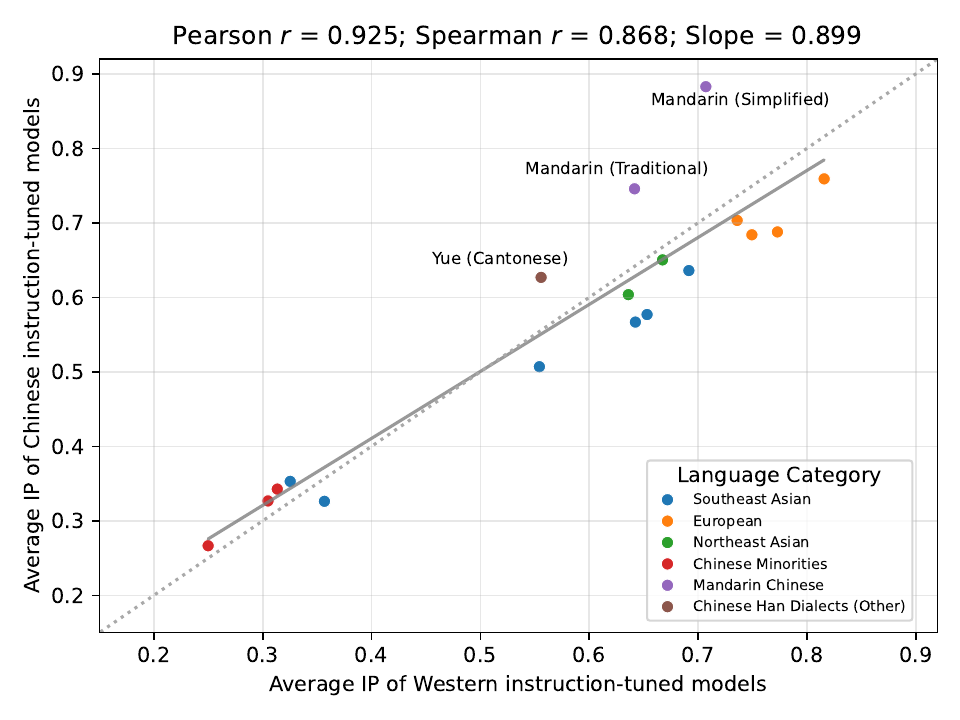}
\caption{Information Parity (IP)}
\label{fig:floresp_ip_scatter_chat}
\end{subfigure}
\hfill
\begin{subfigure}{0.48\textwidth} 
\includegraphics[width=\textwidth]{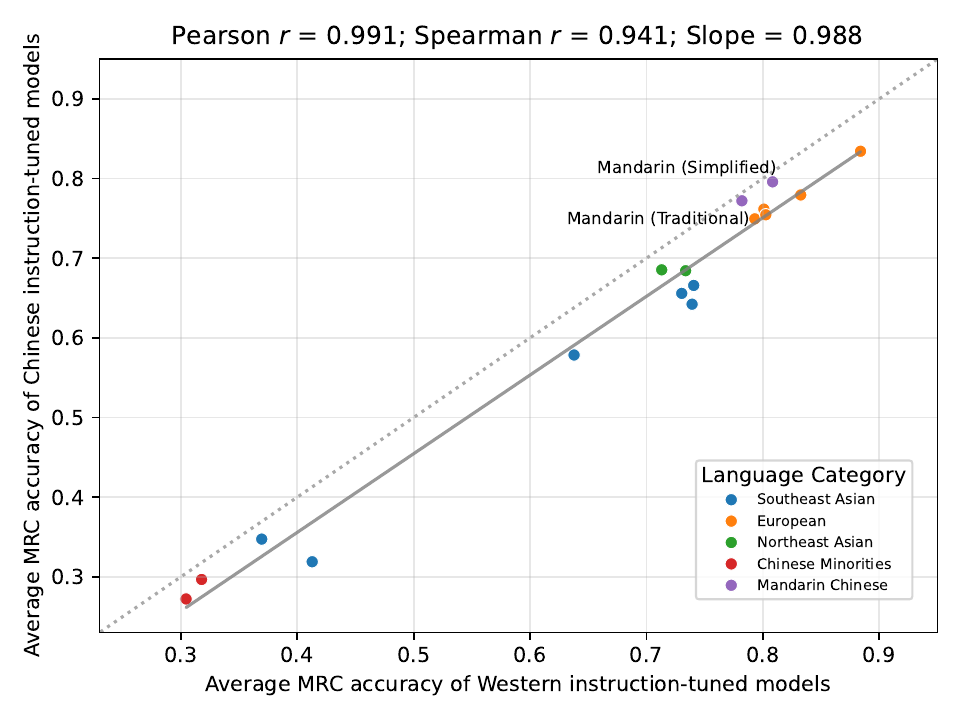}
\caption{{Machine Reading Comprehension (MRC)} accuracy}
\label{fig:bele_accuracy_scatter_chat}
\end{subfigure}
\caption{Performance across languages is highly correlated between the Chinese and Western instruction-tuned models for IP and MRC accuracy. The dotted line is $y=x$ and the solid line is the least squares regression fit. We show the Pearson correlation, Spearman correlation, and the slope of regression fit on top of each plot. All three metrics shows that the two model groups are highly correlated. All correlations are statistically significant ($p<10^{-7}$).}
\label{fig:chat_models_corr}
\end{figure*}
\section{Experiments}
\label{sec:experiments}
\subsection{Models}
We selected our models to optimize for fair comparison while accounting for sufficient variability. In order to provide the fairest comparison among models, we restrict experiments to models of 7--9 billion parameters. This scale is sufficient for at-or-near state of the art performance while limiting computational complexity. We experiment with both base and instruction-tuned open-source LLMs that we access through the Hugging Face \texttt{transformers} library\citep{wolf-etal-2020-transformers}. We selected top-performing models with significant traction.\footnote{Code is available at: \url{https://github.com/andreawwenyi/DoChineseModels_FAccT26}}

We evaluate the following models developed by \textbf{Chinese} organizations (base models in paranthesis): Qwen2.5-7B-Instruct (Qwen2.5-7B) \citep{yang2024qwen2_5}, Yi-1.5-9B-Chat (Yi-1.5-9B) \citep{ai2024yi}, DeepSeek-R1-Distill-Qwen-7B \citep{deepseek-r1}, DeepSeek-R1-Distill-Llama-8B \citep{deepseek-r1}, InternLM3-8b-instruct (InternLM2.5-7B) \citep{cai2024internlm2}, and Baichuan2-7B-Chat (Baichuan2-7B-Base) \citep{yang2023baichuan}. We also evaluate the following \textbf{Western-developed} models: Llama-3-8B-Instruct (Llama3-8B) \citep{llama3modelcard}, Mistral-7B-Instruct-v0.3 (Mistral-7B-v0.3) \citep{jiang2023mistral}, OLMo-2-1124-7B-Instruct (OLMo-2-1124-7B) \citep{olmo20242}, and Gemma-2-9b-it (Gemma2-9b) \citep{gemma_2024}.

\subsection{Data and Experiments}
We selected our benchmark datasets based on several conditional requirements. First, the texts should be created or translated by native speakers to ensure linguistic quality. Second, it must have parallel translations, meaning that the same information is translated into several languages to ensure the same information content in each language. Third, the language coverage should be broad enough to include the languages in our study. 
To the best of our knowledge, only two datasets satisfy these criteria: FLORES+ multilingual machine translation dataset \cite{nllb-24} and the Belebele Parallel Reading Comprehension Dataset \cite{bandarkar-etal-2024-belebele}. 

We add a third dataset specifically developed by Chinese researchers on four Chinese minority languages. The Multilingual Corpus of Minority Languages in China (MC$^2$) \cite{zhang-etal-2024-mc2} is especially valuable for reflecting how Tibetan, Uyghur, Kazakh, and Mongolian as minority languages are used within China specifically. While not a parallel translation dataset, MC$^2$ gives valuable insights and expands coverage of languages underrepresented in the first two datasets.

We design an experiment using each of the three datasets. Together, the three experiments cover both task-agnostic and task-dependent metrics, allowing us to evaluate the LLMs' multilingual performance through a breadth of tasks rather than relying on a single performance metric. We describe the setup for each experiment below.
\paragraph{Experiment 1: Information Parity}
We conduct a task-agnostic evaluation utilizing the FLORES+ benchmark for multilingual machine translation~\citep{goyal-etal-2022-flores, nllb-24}. The benchmark has 997 English samples sentences from Wiki sources.\footnote{We use the \texttt{dev} split of FLORES+.} These sentences are translated into target languages by native speakers and professional translators. This high quality dataset includes parallel translations across around 200 language variants, including many low-resource languages. Example English sentences are in Appendix \ref{apd:flores-sample}.

To measure multilingual capabilities, we calculate Information Parity (IP), an index proposed by \citet{tsvetkov-kipnis-2024-information}. IP aims to measure the comparative efficiency of representing the same information in a reference language $R$ to a target language $L$. Suppose a text input in the reference language $text_R$ is translated to a target language $text_L$, and NLL(text) is the sum of negative log-likelihood of a text input (refer to Appendix \ref{eq:nll} for formal definition), IP is defined by:
\begin{equation}
IP(text_L) = \frac{NLL(text_R)}{NLL(text_L)}
\end{equation}

\begin{figure*}
\begin{subfigure}{0.48\textwidth} 
\includegraphics[width=\textwidth]{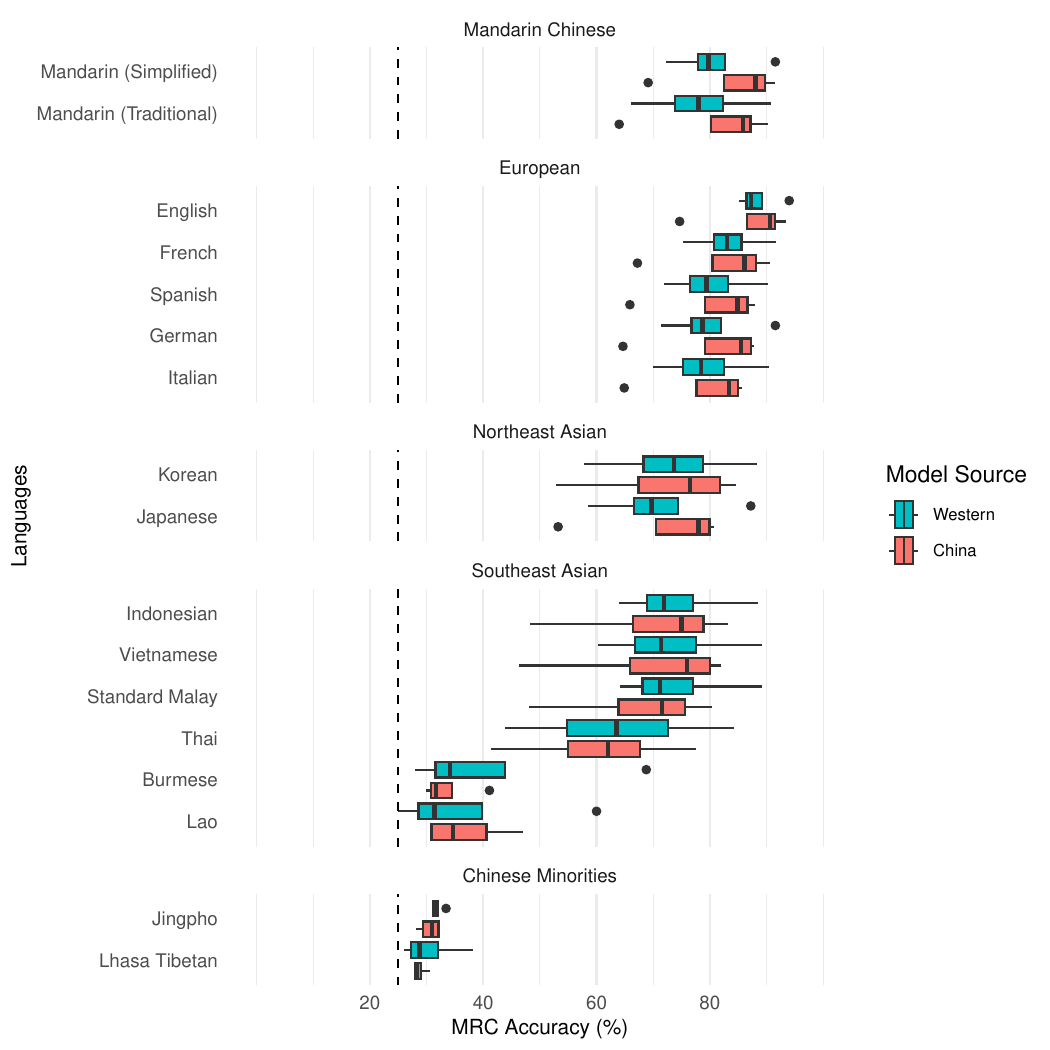}
\caption{Instruction-tuned models}
\label{fig:bele_whisker_chat}
\end{subfigure}
\hfill
\begin{subfigure}{0.48\textwidth} 
\includegraphics[width=\textwidth]{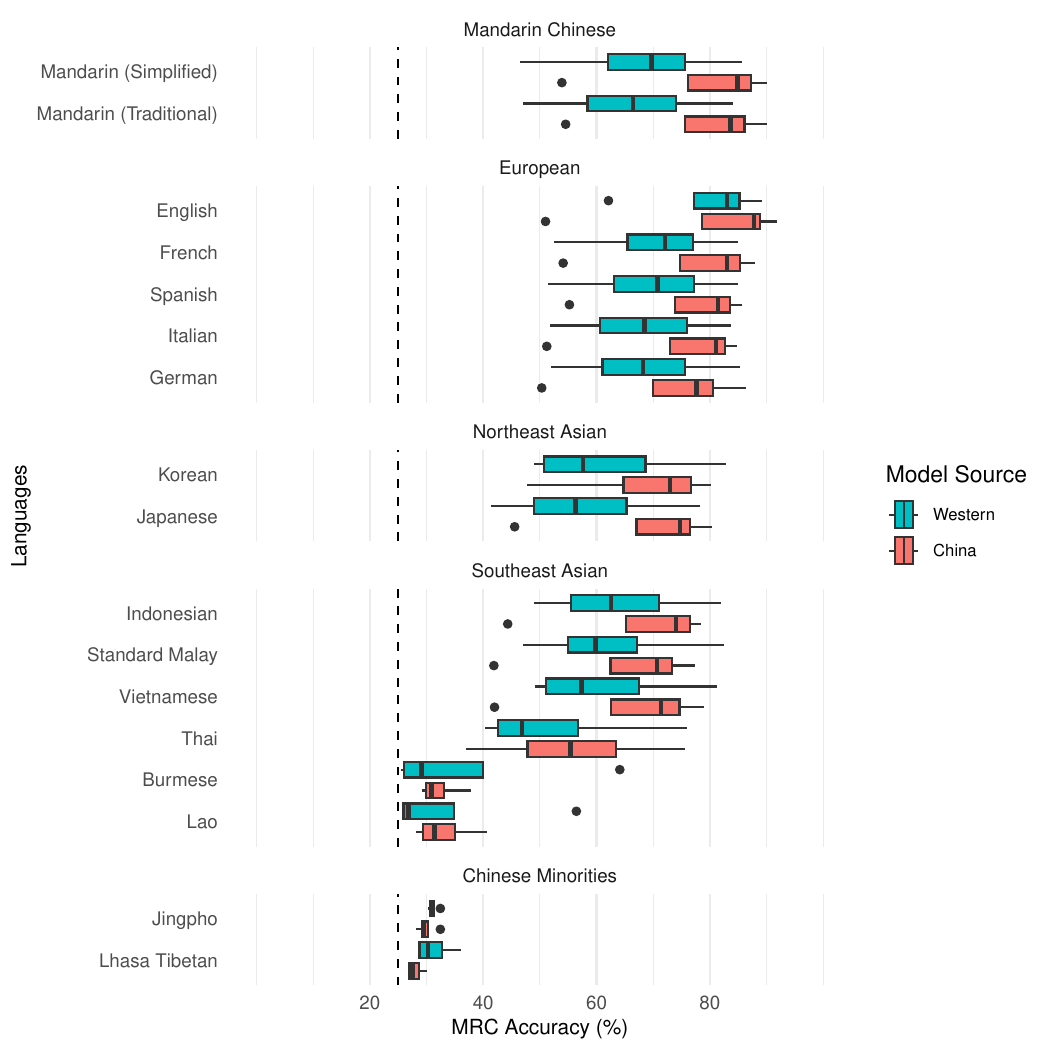}
\caption{Base models}
\label{fig:bele_whisker_base}
\end{subfigure}
\caption{ MRC accuracy of Chinese-developed vs. Western-developed models. Chinese models have higher accuracy in Mandarin reading comprehension questions than their Western counterparts. With base models, Chinese-developed models have higher accuracy than Western-developed models in most languages except Burmese, Lao, Jingpho, and Tibetan. But both groups are similarly performing on instruction-tuned models across all languages.}
\label{fig:bele_whisker}
\end{figure*}
We use English as the reference language for $text_R$ because all models have been shown to be good at it. Higher IP score means higher efficiency in language-agnostic information representation. In other words, language input $text_L$ with higher $IP(text_L$) score means closer alignment with English reference input $text_R$.

Compared to other popular task-agnostic metrics such as tokenization parity \citep{petrov2023language} and fertility \citep{rust-etal-2021-good}, IP is a better predictor of downstream task performance \citep{tsvetkov-kipnis-2024-information}. It is also a more robust multilingual measure than perplexity  because it is less affected by tokenizer differences \citep{wang2022perplexity}. We evaluate IP on 18 language variants from FLORES+, spanning languages spoken in China, Northeast Asia, Southeast Asia, and Europe. See Table \ref{tab:langs} for the full list.

\paragraph{Experiment 2: Machine Reading Comprehension}
We evaluate the models' natural language understanding (NLU) performance on the Belebele benchmark, a multilingual machine reading comprehension (MRC) dataset \citep{bandarkar-etal-2024-belebele}. The dataset comprises of multiple-choice questions. Each question has a passage with four answer choices and one correct answer. There are a total of 900 multiple-choice questions, each translated to 122 high- and low-resource languages. The dataset is curated and verified by translators fluent in both English and the target language. 

We follow the experimental setup in the original Belebele paper to evaluate the models. Specifically, we query models with zero-shot prompts and calculate accuracy of the models’ answers to the multiple-choice reading comprehension questions. We evaluate models on 17 languages from the Belebele dataset, spanning languages spoken in China, Northeast Asia, Southeast Asia, US/Europe (Table \ref{tab:langs}). Refer to Appendix \ref{apd:sec:bele-prompt} for an example passage and prompt.

\paragraph{Experiment 3: Language Identification}
As Chinese minority languages are under-represented in the previous two experiments, we conduct an additional experiment on four minority languages: \textit{Tibetan}, \textit{Mongolian}, \textit{Kazakh}, and \textit{Uyghur}. We use the Multilingual Corpus of Minority Languages in China (MC$^2$) by \citet{zhang-etal-2024-mc2}, which contains web text in these four languages collected from high-quality sources such as news websites. To evaluate whether models can identify these minority languages, we prompt each model to determine the language of 101 texts from the MC$^2$ corpus.\footnote{Kazakh and Mongolian both have multiple writing systems. MC$^2$ collects the writing system predominantly used by speakers in China: Kazakh in Arabic script and Mongolian in Traditional Mongolian script. } We adopt a lenient approach, counting the output as correct if it contains the language name. Refer to Appendix \ref{apd:sec:mc2-prompt} for the prompt.

\begin{figure*}
\begin{subfigure}{0.48\textwidth} 
\includegraphics[width=\textwidth]{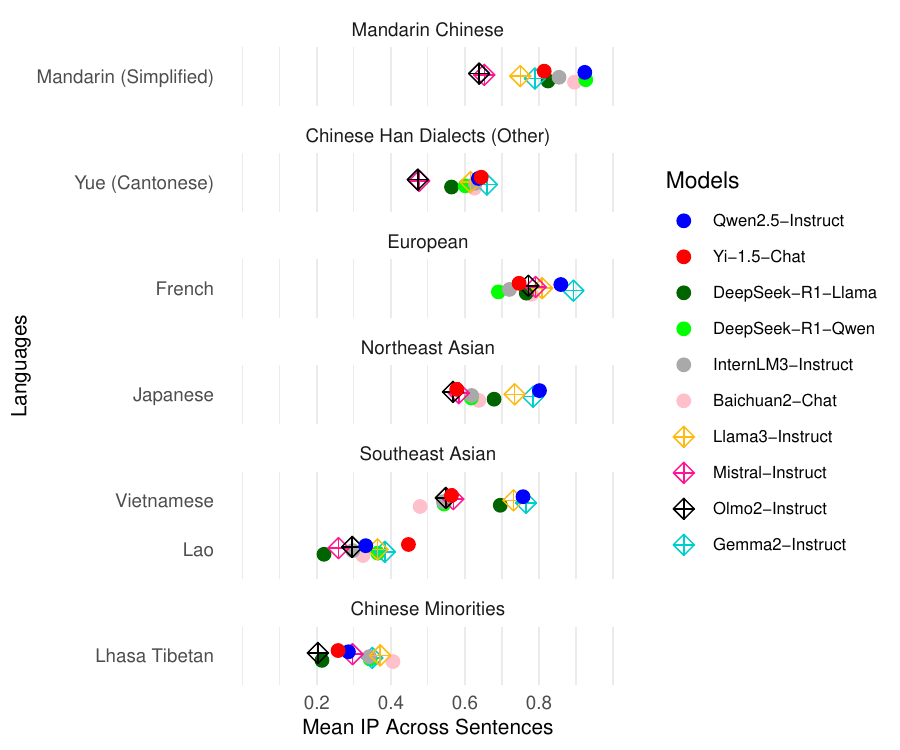}
\caption{Information Parity (IP)}
\label{fig:floresp_ip_boxplot_few_chat_models}
\end{subfigure}
\hfill
\begin{subfigure}{0.48\textwidth} 
\includegraphics[width=\textwidth]{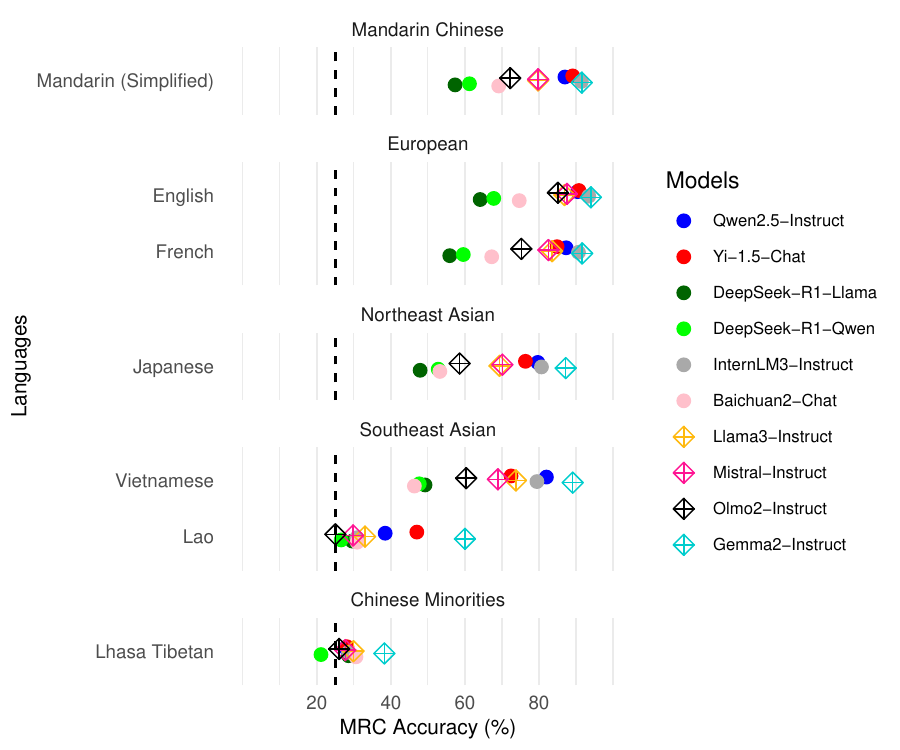}
\caption{MRC Accuracy}
\label{fig:bele_boxplot_few_chat_models}
\end{subfigure}
\caption{Average IP (vs. English) and MRC Accuracy of each instruction-tuned model for select languages. In both figures, Chinese-developed LLMs are represented by circle markers, and Western-developed LLMs by diamond-plus markers. The vertical line in the MRC figure is the $0.25$ random baseline. 
All Chinese-developed LLMs have higher IP than their Western counterparts in Simplified Mandarin. In MRC accuracy, \texttt{Gemma2-Instruct} is consistently the highest, and \texttt{DeepSeek} models underperform. The order of models stays similar across languages, except in Tibetan, where most models are near random.}
\label{fig:}
\end{figure*}
\section{Results} \label{sec:results}
\subsection{Experiment 1: Information Parity (IP) on FLORES+ Dataset}
We show instruction-tuned models' IP scores averaging across 997 translated sentences on a selection of languages (Figure \ref{fig:floresp_ip_boxplot_few_chat_models}). \texttt{Gemma2} and \texttt{Qwen2.5} have the highest average IPs in most languages. In Simplified Mandarin, all Chinese-developed models have higher IP than their Western counterparts. See Appendix Figure \ref{fig:floresp_ip_boxplot_full} for the full model-language results.

The IP distribution of Chinese- and Western-developed models on all 18 languages is shown in Figure \ref{fig:floresp_ip_whisker_full}. We exclude \texttt{DeepSeek-R1-Llama} from this figure as it is a US model (Llama) fine-tuned by a Chinese organization (Deepseek). We observe that Chinese LLMs, both base and instruction-tuned, have significantly higher IPs than Western-developed models in Mandarin Chinese. In Cantonese as well, Chinese LLMs have higher IPs and lower variance. Chinese instruction-tuned models have lower IP variability in Japanese, Korean, Vietnamese, and Malay than base models and Western-developed instruction-tuned models have higher IPs in European language. Despite these performance differences, when it comes to Chinese minority languages, both Chinese-developed and Western-developed LLMs have low IP and similar variance. 
See Appendix Figure \ref{fig:floresp_instruction-tuned-effect} for full results of instruction-tuned models.

We find high correlations between Chinese and Western-developed models' average IP across languages. We assess the correlation through commonly-used statistical measures for correlation: Pearson and Spearman correlations. We also report the slope of the least-squares linear regression line. 
For instruction-tuned models, the Pearson correlation is $0.925$, the Spearman correlation is $0.868$, and the slope of the least-squares regression line is $0.899$ (Figure \ref{fig:floresp_ip_scatter_chat}). For the base models, the Pearson correlation is $0.929$, Spearman is $0.889$, and slope is $0.968$ (Appendix Figure \ref{fig:floresp_ip_scatter_base}). All slopes are close to $y=x$, and all correlations are statistically significant ($p<10^{-7}$).

\subsection{Experiment 2: Machine Reading Comprehension (MRC) on Belebele Dataset}

The MRC accuracy for all models on selected languages shows a wider distribution than IP (Figure \ref{fig:bele_boxplot_few_chat_models}). We see \texttt{Gemma2-Instruct} consistently outperforms other models, whereas \texttt{Deepseek-R1-Qwen} and \texttt{Baichuan2-Chat} consistently underperform (See Appendix Figure \ref{fig:bele_boxplot_full} for full language/model breakdown for both instruction-tuned and base models). 

We compare Chinese-developed v.s. Western-developed LLMs MRC accuracies across 17 languages in Figure \ref{fig:bele_whisker}.\footnote{As in Experiment 1, we exclude \texttt{DeepSeek-R1-Llama}. We additionally exclude \texttt{DeepSeek-R1-Qwen} from Figure \ref{fig:bele_whisker_chat}, as we find the model to be particularly sensitive to the choice of chat template (see Appendix Figure \ref{fig:chat-template-effect-MRC}), unlike other models. For comparability, we apply a consistent prompt across all models in the figure; therefore, we omit \texttt{DeepSeek-R1-Qwen} from the grouped bar results in this figure. We instead report its performance using a more effective chat template in Appendix Figure \ref{fig:bele_whisker_full_chat_models_deepseekQwen}.} As in Experiment 1, Chinese LLMs are significantly better in Mandarin Chinese. MRC accuracies in other languages are similar among instruction-tuned models. However, with base models, Chinese-developed models are better than Western-developed ones in almost every language except in those with low accuracy, namely Burmese, Lao, Jingpho and Tibetan (Figure \ref{fig:bele_whisker_base}). 

\begin{figure*}[t]
  \begin{subfigure}{0.48\textwidth}  
        \centering
        \includegraphics[width=\textwidth]{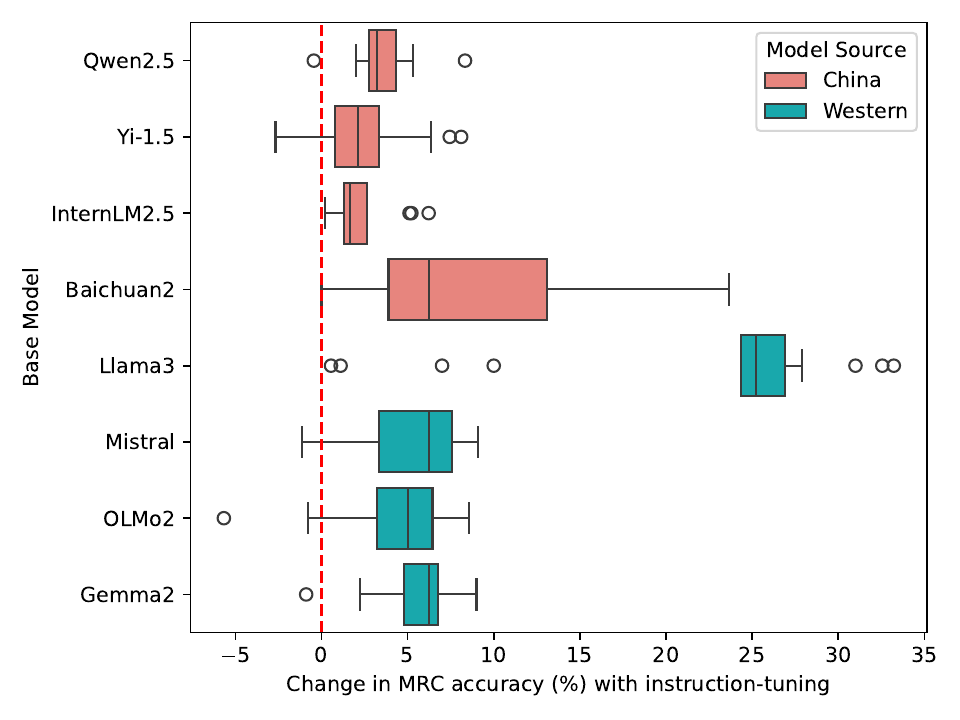}
    \end{subfigure}
    \hfill
    \begin{subfigure}{0.48\textwidth}
        \centering
        \includegraphics[width=\textwidth]{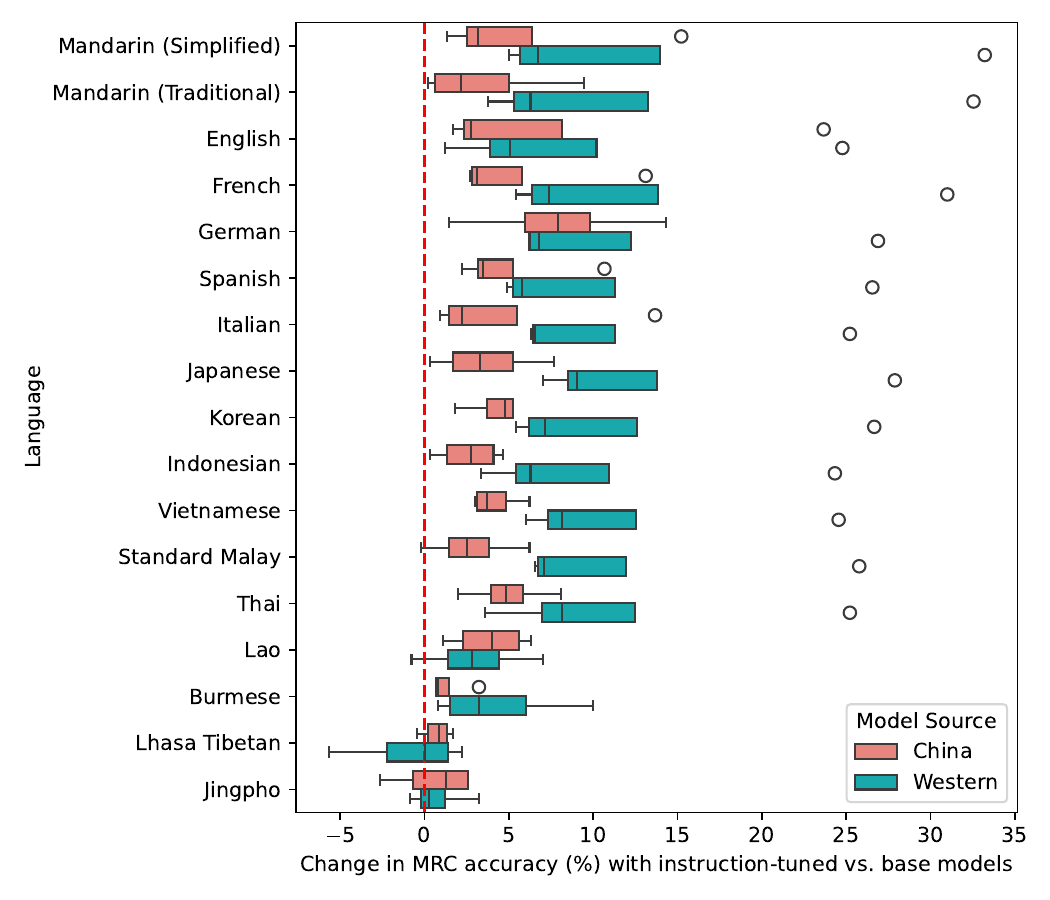}
    \end{subfigure}
    \caption{Change in MRC accuracy with instruction-tuned vs. base models. Instruction-tuned models generally outperform their base versions, especially \texttt{Llama3}. Compared to Chinese-developed models, Western-developed instruction-tuned models show a larger accuracy gain over their base models across most languages.}
    \label{fig:instruction-model-effect-MRC}
\end{figure*}

We observe that Chinese-developed and Western-developed models exhibit different results when instruction-tuned. As expected, we find that instruction-tuned models have higher accuracy than their base counterparts overall (Figure \ref{fig:instruction-model-effect-MRC}). However, this effect is larger in Western-developed models across most languages except in Burmese, Lao, Jingpho and Tibetan. Notably, the instruction-tuning effect on \texttt{Llama3} far exceeds others' --- twice as big as the other base types --- with the median increase of 25\% over the base.  While we tested various chat templates and system prompt designs for instuction-tuned models, we do not observe significant differences except for \texttt{DeepSeek} models. See Appendix Figure \ref{fig:chat-template-effect-MRC}.

With MRC, the correlation between Chinese-developed and Western-developed models is even stronger than seen in Experiment 1. For instruction-tuned models (Figure \ref{fig:bele_accuracy_scatter_chat}), the Pearson correlation is $0.991$, the Spearman correlation is $0.941$, and the slope of least-squares regression line is $0.988$. For base models (Appendix Figure \ref{fig:bele_accuracy_scatter_base}), the Pearson correlation is $0.984$, Spearman is $0.897$ and the slope is $1.21$. All correlations are statistically significant ($p<10^{-7}$) and all regression lines are close to $y=x$.

\subsection{Experiment 3: Language Identification on Chinese Minority Languages}

\begin{wrapfigure}{R}{0.5\textwidth}
\centering
\includegraphics[width=0.48\textwidth]{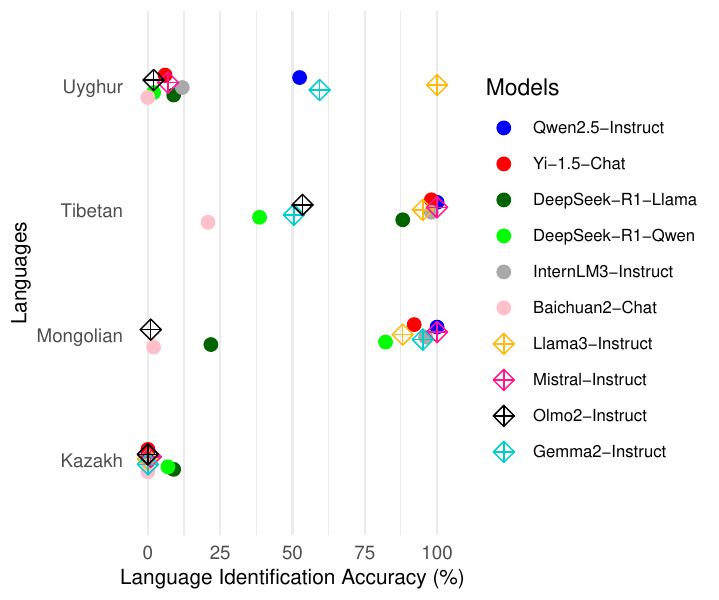}
\caption{Language identification accuracy for four Chinese minority languages in the MC$^2$ corpus.}
\label{fig:mc2_langpred_boxplot_chat_models}
\end{wrapfigure}

Our language identification experiments on four Chinese minority languages reveals that overall, models have higher success identifying Tibetan and Mongolian than Uyghur and Kazakh (Figure \ref{fig:mc2_langpred_boxplot_chat_models}). All models are bad at identifying Kazakh. Most models cannot identify Uyghur. And most models can identify at least 75\% of Tibetan and Mongolian. \texttt{Llama3-Instruct} is uniquely good at Uyghur, and is also top-performing in Tibetan and Mongolian. \texttt{Baichuan2} is consistently the worst in every language. For results of base models, see Appendix Figure \ref{fig:mc2_langpred_boxplot_base_models}.

\section{Discussion} \label{sec:discussion}
We make the following three observations from our experiments.
First, Chinese-developed models are better at Mandarin than are Western-developed models. Second, Chinese models are just as bad at Chinese minority languages as are Western-developed models. Some Chinese models fail to recognize minority languages such as Uyghur and Kazakh. Third, Chinese models' performance across languages highly correlates with those of Western-developed models. Like their Western counterparts, Chinese-developed models are better at European languages such as French and German, but underperform at Chinese minority languages.
Overall,  our results support the \textbf{Mandarin Hypothesis}---Chinese models are better at Mandarin than Western-developed models but not at other languages spoken in China (Fig  \ref{fig:floresp_ip_whisker_full}, \ref{fig:bele_whisker}). Meanwhile, we cannot reject the \textbf{Null Hypothesis}, as we observe a high correlation on multilingual performance between Western and Chinese-developed models (Fig \ref{fig:chat_models_corr}). 

Our mandarin hypothesis findings suggest that the de facto practice for language AI development in China appears to be Mandarin-first. Structurally, there are no observable signs of investment in minority language technology development in China. Chinese model developers may not have easily accessible resources for Chinese minority language research, nor are there obvious incentives to meet performance standards in local languages other than Mandarin. This is in contrast to China's mid-twentieth century linguistic pluralism approach, where the Chinese government following the Communist revolution invested resources to collect, categorize, and study minority languages in China \citep{mullaney2011coming}. Instead, it aligns with PRC's more recent assimilationist policy aiming to ``promote integration across ethnic groups'' by establishing Mandarin as the unifying language of China. The recently passed ethnic minority law, called ``Promoting Ethnic Unity and Progress,'' not only promotes Mandarin as the language for education and government, but mandates that Mandarin be given ``prominence in placement'' when used alongside minority languages in public settings \cite{2026-ChinaPassesNew}.

On the other hand, our null hypothesis findings suggest that there \textit{is} observable pressure for Chinese open-weight LLMs developers to optimize for global benchmark performance. Preliminary evidence from selected technical reports further points to the prominence of such benchmarks: Chinese-developed open-weight LLMs are largely evaluated against Western AI standards. For example, \texttt{DeepSeek-R1} is evaluated on ten English benchmarks, compared to three in Mandarin Chinese. The reasoning-related benchmarks that  \texttt{DeepSeek-R1} evaluates itself against are also in English \citep{deepseek-r1}. The \texttt{Qwen2.5} family is explicitly assessed on multiple languages, including Arabic, Japanese, Korean, and Turkish, but not on Chinese minority languages \citep{yang2024qwen2_5}. This observation contributes to the body of work examining how a small number of institutions set benchmarking practices that have undue influence over AI systems, and in turn on the greater public's cultural and linguistic norms \cite{birhane2020-AlgorithmicColonizationAfrica, agarwal2025-AISuggestionsHomogenize, tao2024-CulturalBiasCultural,varshney2025-DecolonialAIAlignment,mirowski2024-RobotWalksBar, koch2021-ReducedReusedRecycled}. 

Further work is needed to systematically analyze evaluation practices and examine the structural and institutional factors shaping development priorities, such as state policy incentives, funding structures, and participation in global benchmarking ecosystems. Such analyses would likely require qualitative approaches (e.g., interviews, document analysis, and ethnography) that are beyond the scope of this computational study. We hope to see future work engaging with user interviews and other qualitative assessments for further analysis.

\paragraph{Long term implications of homogenization} 
In contrast to the recent trend towards sovereign AI \cite{Alduhishy2024-SovereignAIWhat}, our results reveal a pattern of homogenization in multilingual performance across the models we study, even when model developers operate in distinct linguistic, political, and cultural contexts. This suggests that there are certain normalizing pressures applied on LLM training via global benchmarking practices and shared resources such as training and evaluation data.
The homogenization has several implications. First, we are pushed to ask, to what extent are we training near-identical models at the cost of environmental damage? A growing body of work has discussed the environmental impact of training and deploying generative models \cite{dodge2022-MeasuringCarbonIntensity, rakova2023-AlgorithmsSocialEcologicalTechnologicalSystems, varoquaux2025-HypeSustainabilityPrice,crawford2021-AtlasAIPower}. As countries and entities race to train their own models that don't lead to substantial differences in utility, concerns about environmental impact become even more salient. 

Second, technological homogenization also influences users. Popular technological products themselves apply normalizing pressure on consumers or recipients of technological content. One domain is education. As students are exposed to and utilizing more AI products for learning purposes, AI outputs shape their values, perspectives, and even habits. Homogenized language models with selective linguistic capacity can force users to learn more AI-aligned languages and neglect their own. While today's models may have this effect inadvertently, language support agenda as a political tool of control has historical origins in China and elsewhere.

Third, our observations on homogenization echo the persistent tension between model work and data work in AI development \citep{Sambasivan2021Everyone}. While improvements in overall performance across languages might be the result of algorithmic improvement, relative differences between languages are more sensitive to decisions about data. This suggests opportunities for model developers with access to domain- or language-specific data to differentiate their models and contribute to the diversity of LLMs. For example, Chinese developers have access to Chinese local languages that are otherwise difficult to collect or process by others. Chinese developers and researchers therefore have a unique advantage in tackling the lagging performance of Chinese minority languages by investing in the digitization and evaluation of minority languages in mainland China. In this way, Chinese model developers also have a singular advantage in setting new standards and metrics for its minority languages. 

\section{Limitations and Future Work} \label{sec:conclusion}
We acknowledge several limitations in our work.
First, the models we evaluate on are open-weight, with 7--9 billion parameters, and released on Hugging Face, which could represent a bias in model selection that are not representative of China's LLMs. We note that China has their own Hugging Face equivalent, such as ModelScope and OpenCSG.\footnote{https://www.modelscope.cn/}\footnote{https://opencsg.com/} It is possible that models released on these platforms may be different from the ones on Hugging Face. However, at the time of writing, we find that the popular models on these platforms are also on Hugging Face. We also note that there are other closed-source LLMs developed and widely used in mainland China (e.g. Baidu's ERNIE bot) that may have different performance outputs. However, closed-source models do not give us the level of liberal access we needed for our experiments. 
Second, while we evaluate Chinese and other Asian languages, our experiments only feature benchmarks that have been translated from English. This may introduce biases by way of cultural representation. However, we note a general lack of parallel dataset that is translated from Mandarin. Third, the benchmarks we use are open-source and well-known, possibly exposing them for developers to train or finetune their models on. Fourth, due to data availability, we do not include most of the Chinese Han dialects. In addition to minority languages, China also has significant populations speaking 8--10 different Chinese Han dialects, such as Cantonese, Hokkien, or Shanghainese, in their day-to-day lives. The difficulty in evaluating Han dialects is that, except for Cantonese in Hong Kong, these dialects often do not have standardized writing forms. We also note the relative few number of minority languages we evalutae in this study. In general, data collection efforts are required to evaluate performance on more dialects and minority languages. 

\section*{Generative AI Disclosure Statement}
All text in this paper was originally written by the authors. We used ChatGPT to lightly edit the writing, including to check grammar and to suggest ways to improve the fluency of the writing. We also searched for relevant papers to review using Ai2 Asta. 
\begin{acks}
  We would like to thank Lu Jia Lin, Tiezhen Wang, Anna Choi, Greg Yauney, Rosamond Thalken, Federica Bologna, Shengqi Zhu, Sil Hamilton, Axel Bax, Kiara Liu, Rebecca Hicke, Matthew Wilkens and the anonymous reviewers for their thoughtful feedback. This work received R\&D support from South Korean Deeptech TIPS funds.
\end{acks}

\bibliographystyle{ACM-Reference-Format}
\bibliography{1-reference,2-zotero-references}

\appendix

\section{Sample English sentences from FLORES+ dataset}\label{apd:flores-sample}
\begin{quote}
\textit{On Monday, scientists from the Stanford University School of Medicine announced the invention of a new diagnostic tool that can sort cells by type: a tiny printable chip that can be manufactured using standard inkjet printers for possibly about one U.S. cent each.}
\end{quote}
\begin{quote}
    \textit{``Panama Papers'' is an umbrella term for roughly ten million documents from Panamanian law firm Mossack Fonseca, leaked to the press in spring 2016.}
\end{quote}

\section{Negative Log Likelihood}
\label{eq:nll}
The sum of negative log likelihood of a given text $(t_1, t_2, ..., t_n)$, where $t_i$ is the $i^{th}$ token of the text, under a language model $M$, is defined as followed
\begin{equation}
NLL(text) = NLL(t_1, t_2, ..., t_n) = \sum_{i=1}^n -\log P_M(w_i|w_{1:i-1})
\end{equation}
where $P_M$ is the probability assigned by model $M$. 

\section{Prompt used in Machine Reading Comprehension (Experiment 2)}
\label{apd:sec:bele-prompt}
Below is the prompt format we use for experiment 2 with a question from the Belebele dataset. We construct the prompts following the original paper \citep{bandarkar-etal-2024-belebele}:

\begin{lstlisting}[frame=single,breaklines=true]
Given the following passage, query, and answer choices, output the letter corresponding to the correct answer.
###
Passage:
With the change from the quarter to the half mile run, speed becomes of much less importance and endurance becomes an absolute necessity. Of course a first-class half-miler, a man who can beat two minutes, must be possessed of a fair amount of speed, but endurance must be cultivated at all hazards. Some cross country running during the winter, combined with gymnasium work for the upper part of the body, is the best preparation for the running season.
###
Query:
According to the passage, which of the following would be the most beneficial for a runner preparing for the upcoming season?
###
Choices:
(A) Practicing cross country running in the summer
(B) Focusing on cultivating speed while training
(C) Beating a three minute time
(D) Utilizing the gym to work out the upper body
###
Answer:
\end{lstlisting}

\section{Prompt used in Minority Language Identification (Experiment 3)}
\label{apd:sec:mc2-prompt}
\begin{CJK*}{UTF8}{mj}
\begin{lstlisting}[frame=single,breaklines=true,escapechar=\#]
Identify the language of the given text. Output the English name of the language. Be concise.
    Example:

    Text: #地元メディアの報道によると、空港の消防車が対応中に横転したということです。#
    Language: Japanese

    Text: #그# #조종사는# #비행# #중대장# #딜로크리트# #패타비로# #확인되었다#.
    Language: Korean

    Text: <input text>
    Language: 
\end{lstlisting}
\end{CJK*}

\begin{figure*}[ht]
  \begin{subfigure}{0.48\textwidth}  
        \centering
        \includegraphics[width=\textwidth]{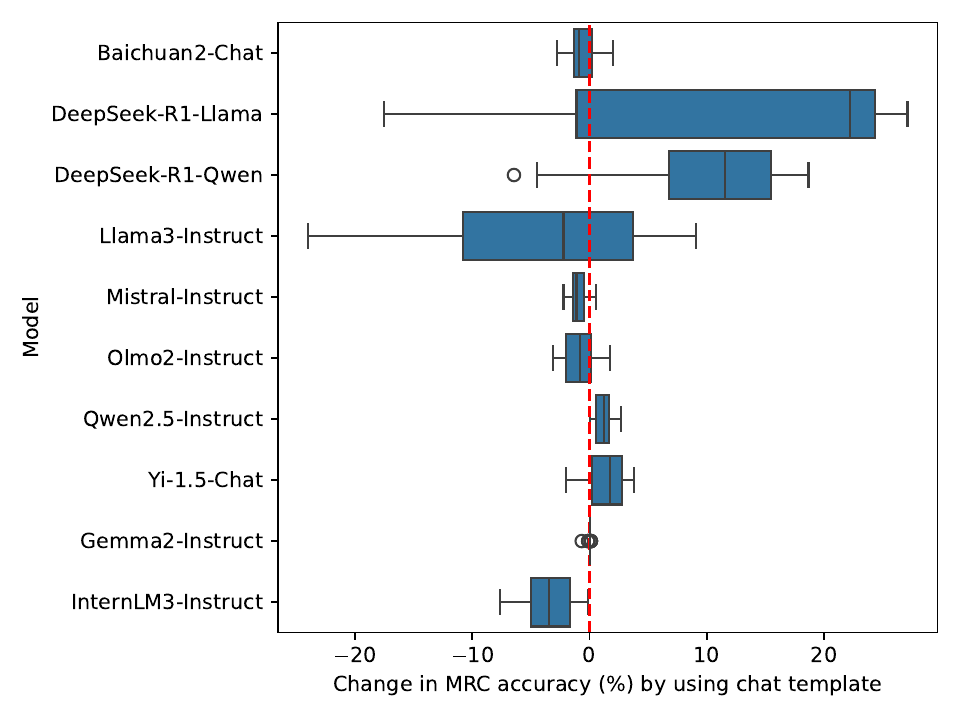}
    \end{subfigure}
    \hfill
    \begin{subfigure}{0.48\textwidth}
        \centering
        \includegraphics[width=\textwidth]{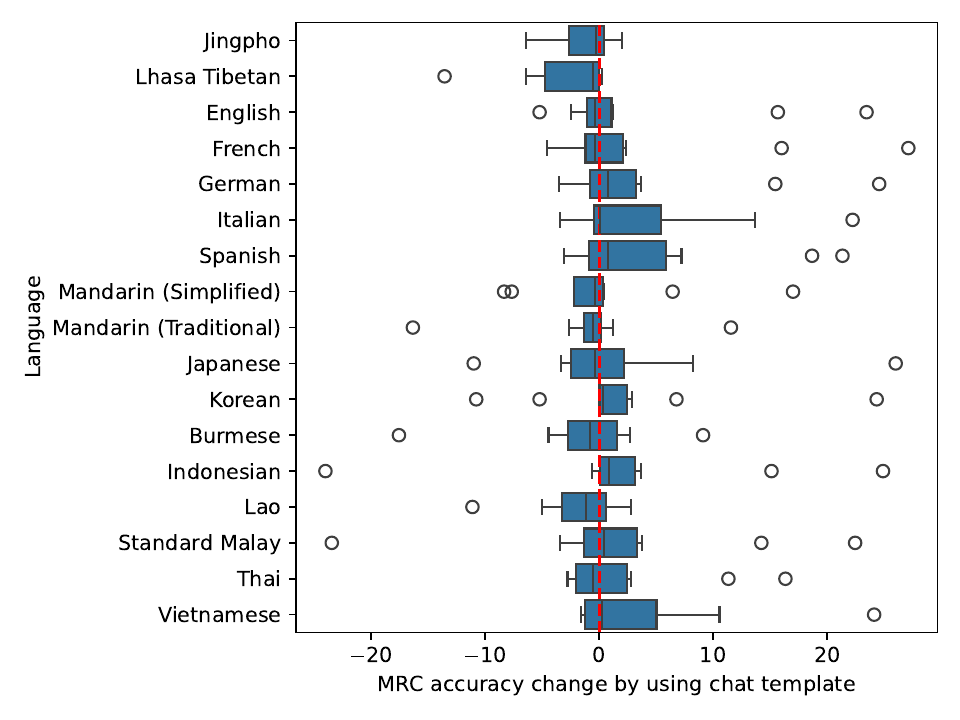}
    \end{subfigure}
    \caption{Change in MRC accuracy by using chat template. Overall, using chat template and system prompt does not significant influence model performance on MRC, with the exception for \texttt{DeepSeek} models. We observe that custom chat templates better activate the ``$<$think$>$'' pattern in DeepSeek models.}
    \label{fig:chat-template-effect-MRC}
\end{figure*}

\begin{figure*}
\begin{subfigure}{0.48\textwidth} 
\includegraphics[width=\textwidth]{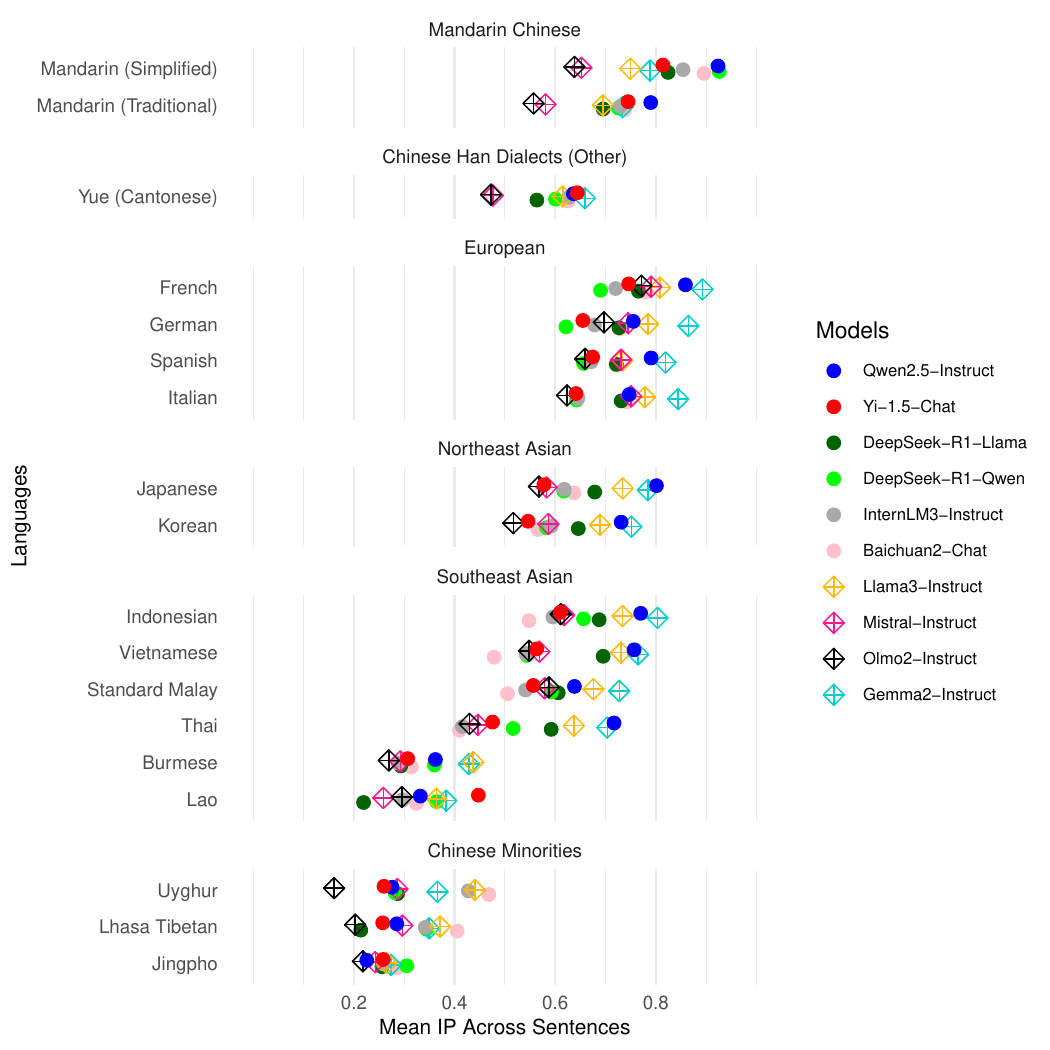}
\caption{Instruction-tuned models}
\label{fig:floresp_ip_boxplot_full_chat}
\end{subfigure}
\hfill
\begin{subfigure}{0.48\textwidth} 
\includegraphics[width=\textwidth]{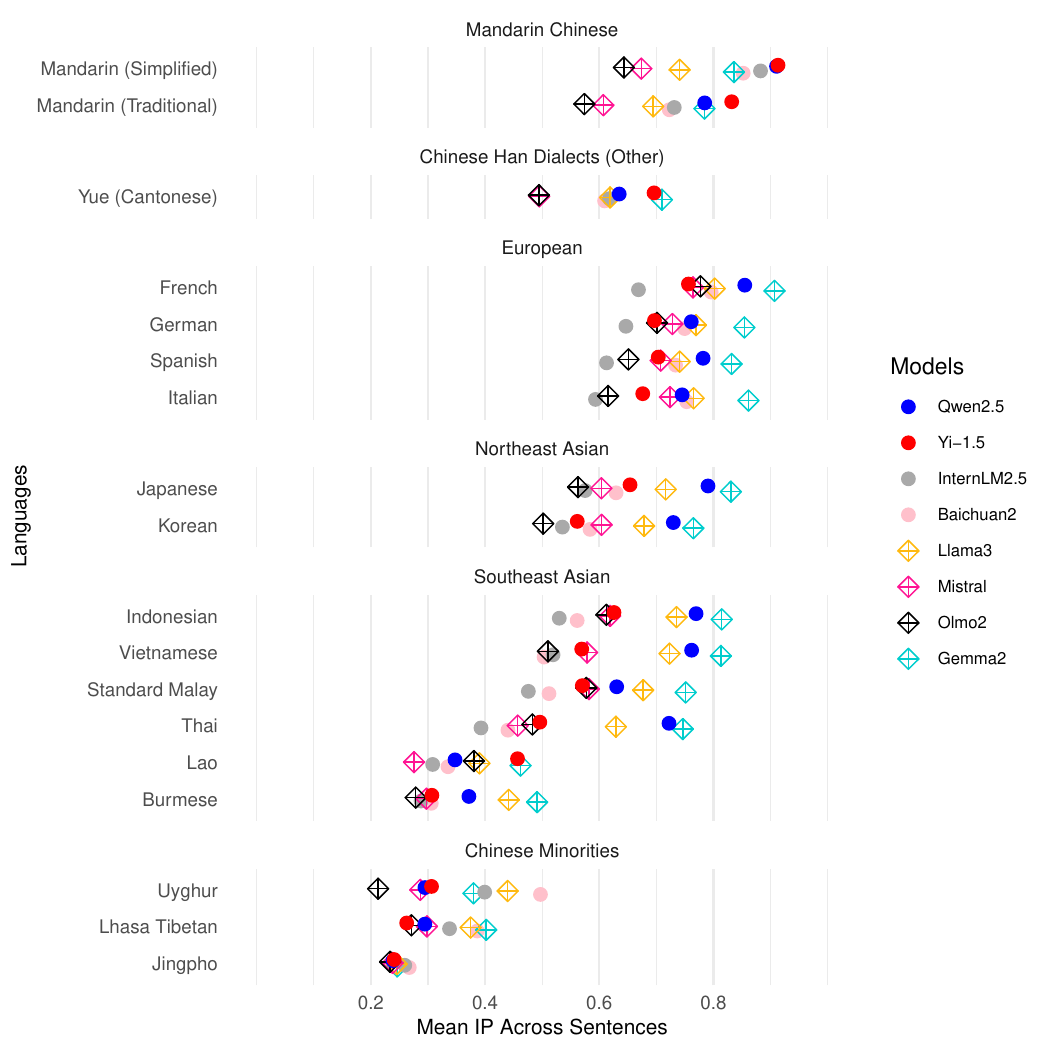}
\caption{Base models}
\label{fig:floresp_ip_boxplot_full_base}
\end{subfigure}
\caption{Experiment 1 full language and models breakdown.}
\label{fig:floresp_ip_boxplot_full}
\end{figure*}

\begin{figure*}
\begin{subfigure}{0.48\textwidth} 
\includegraphics[width=\textwidth]{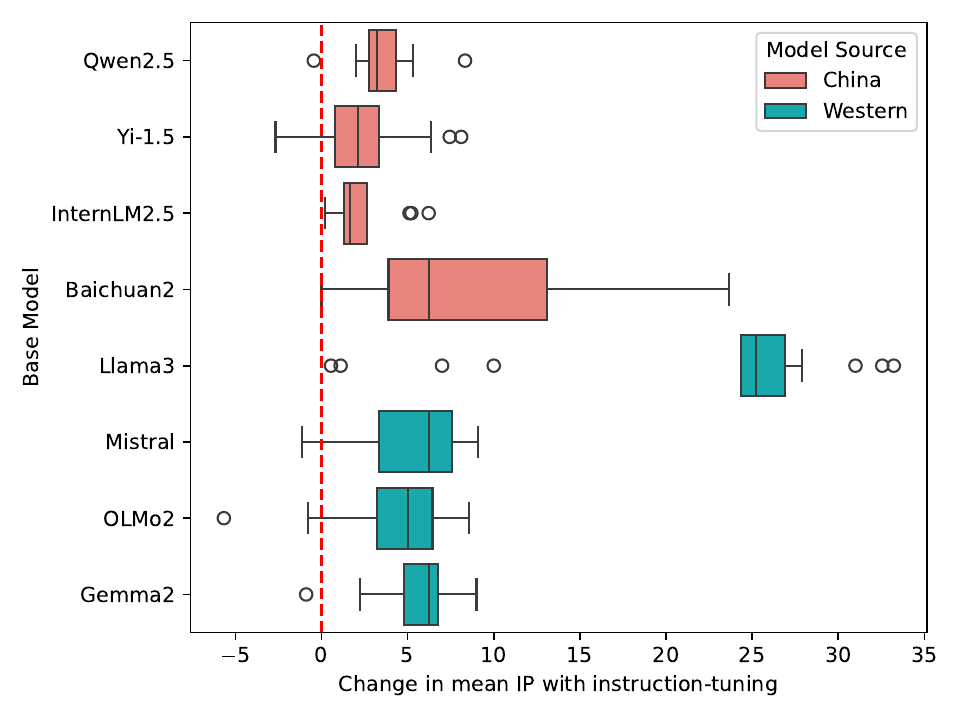}
\end{subfigure}
\hfill
\begin{subfigure}{0.48\textwidth} 
\includegraphics[width=\textwidth]{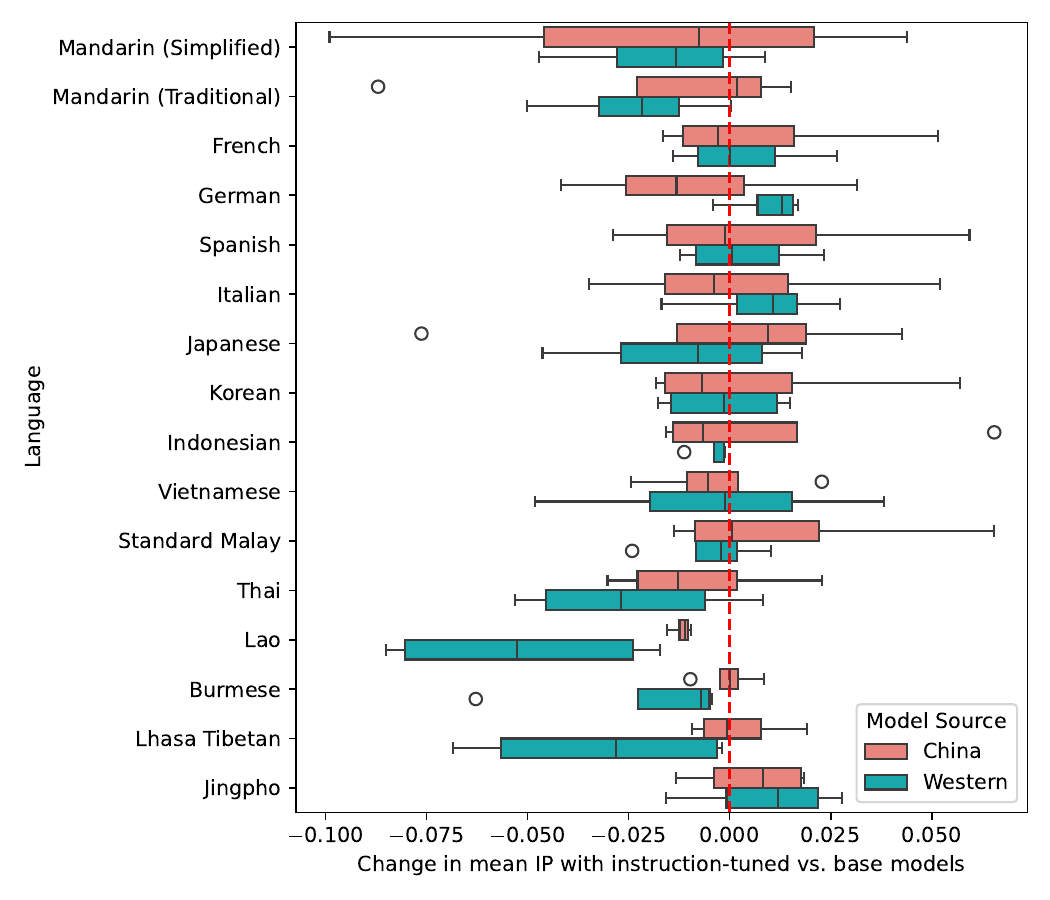}
\end{subfigure}
\caption{Change in average IP with instruction-tuned vs. base models}
\label{fig:floresp_instruction-tuned-effect}
\end{figure*}

\begin{figure*}
\begin{subfigure}{0.48\textwidth} 
\includegraphics[width=\textwidth]{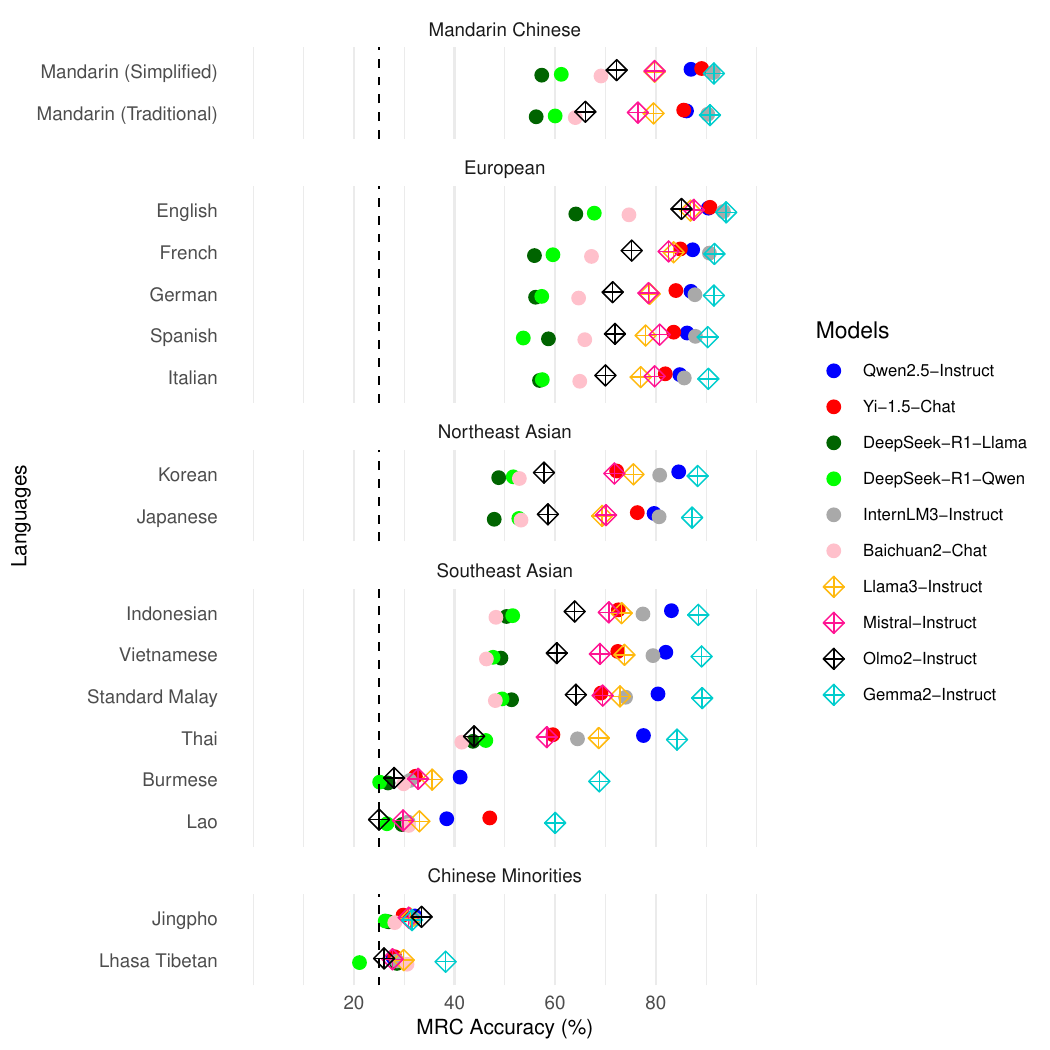}
\caption{Instruction-tuned models}
\label{fig:bele_boxplot_full_chat}
\end{subfigure}
\hfill
\begin{subfigure}{0.48\textwidth} 
\includegraphics[width=\textwidth]{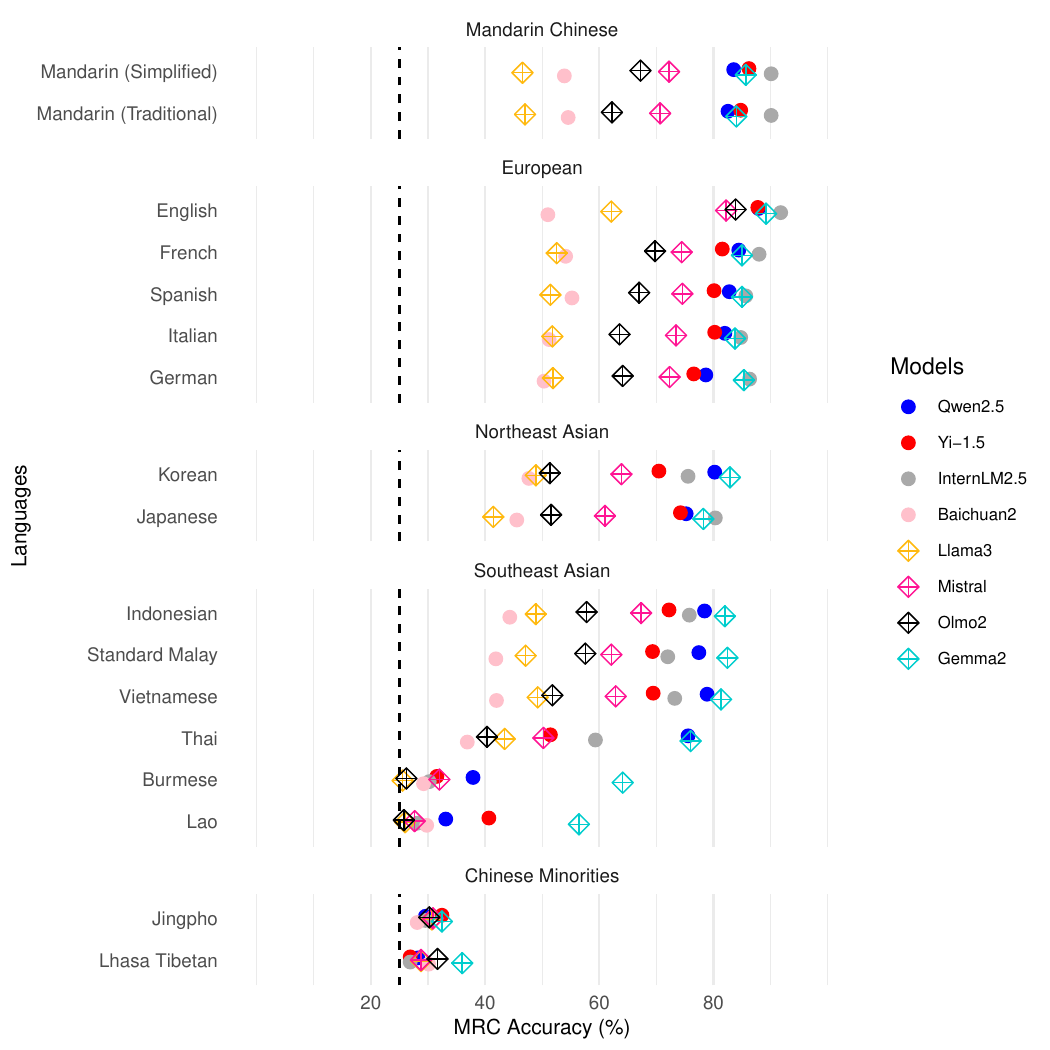}
\caption{Base models}
\label{fig:bele_boxplot_full_base}
\end{subfigure}
\caption{MRC Accuracy of instruction-tuned models (a) and base models (b) on 17 languages. The ranking of model performance are largely
 consistent across languages, except for Burmese, Lao, Jingpho, and Tibetan. \texttt{Gemma2}, \texttt{IntermLM2.5} and \texttt{Qwen2.5} top the charts among both instruction-tuned and base models. \texttt{Gemma2} significantly outperforms other models in Burmese and Lao. Models perform similarly poor, around random baseline of 0.25, in Chinese minorities languages.}
\label{fig:bele_boxplot_full}
\end{figure*}

\begin{figure*}
\begin{subfigure}{0.5\textwidth} 
\includegraphics[width=\textwidth]{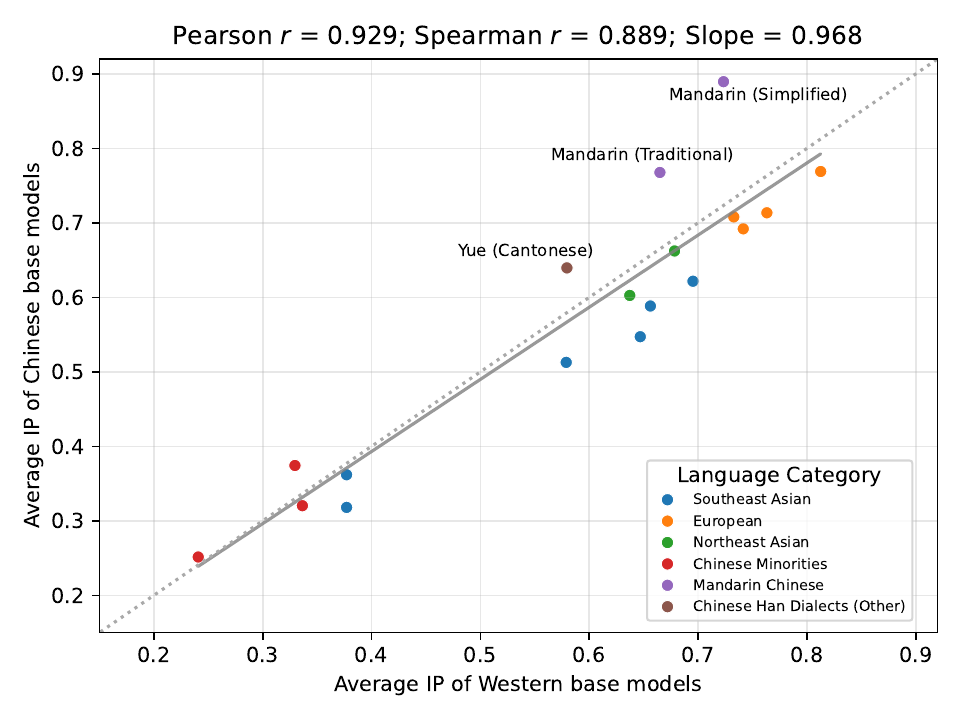}
\caption{IP}
\label{fig:floresp_ip_scatter_base}
\end{subfigure}
\hfill
\begin{subfigure}{0.48\textwidth} 
\includegraphics[width=\textwidth]{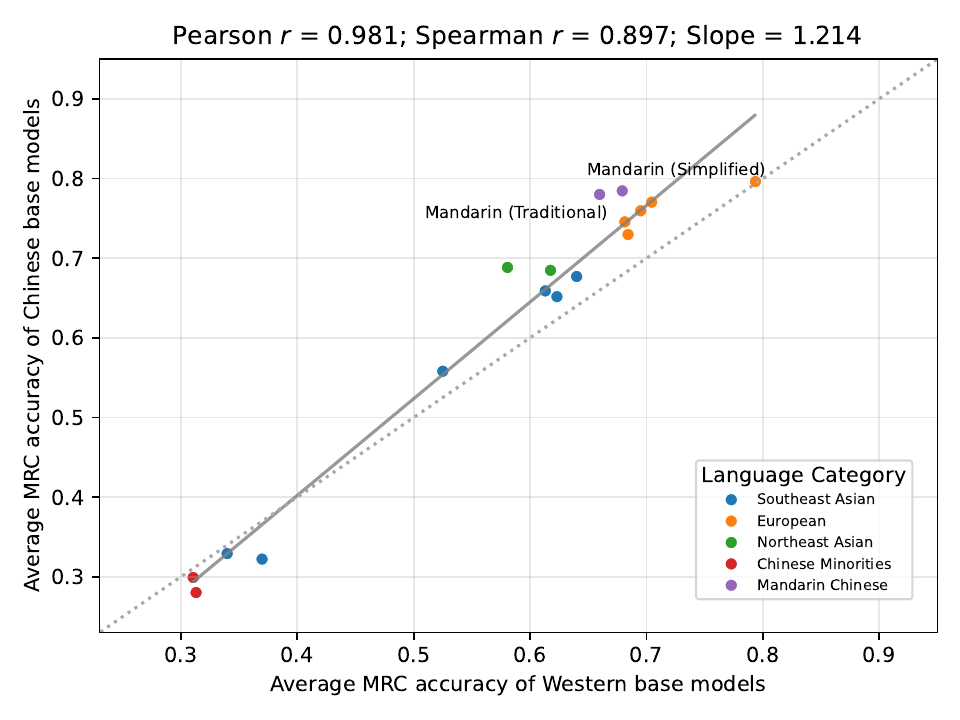}
\caption{MRC Accuracy}
\label{fig:bele_accuracy_scatter_base}
\end{subfigure}
\caption{Correlation of IP and MRC accuracy between Chinese and Western base models. Across languages, the two model groups have a Pearson correlation of $0.929$ in IP and $0.981$ in MRC accuracy. The slopes of fitted line are $0.97$ (IP) and $1.21$ (MRC).}
\label{fig:base_model_corr}
\end{figure*}

\begin{figure*}
    \centering
\includegraphics[width=0.5\textwidth]{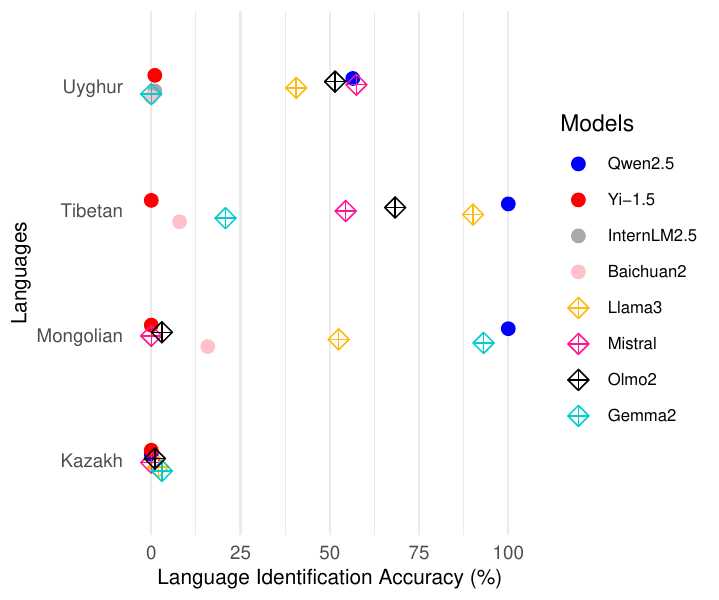}
\caption{Language identification accuracy of base models on MC$^2$ data.}
\label{fig:mc2_langpred_boxplot_base_models}
\end{figure*}

\begin{figure*}[t]
        \centering
        \includegraphics[width=0.5\textwidth]{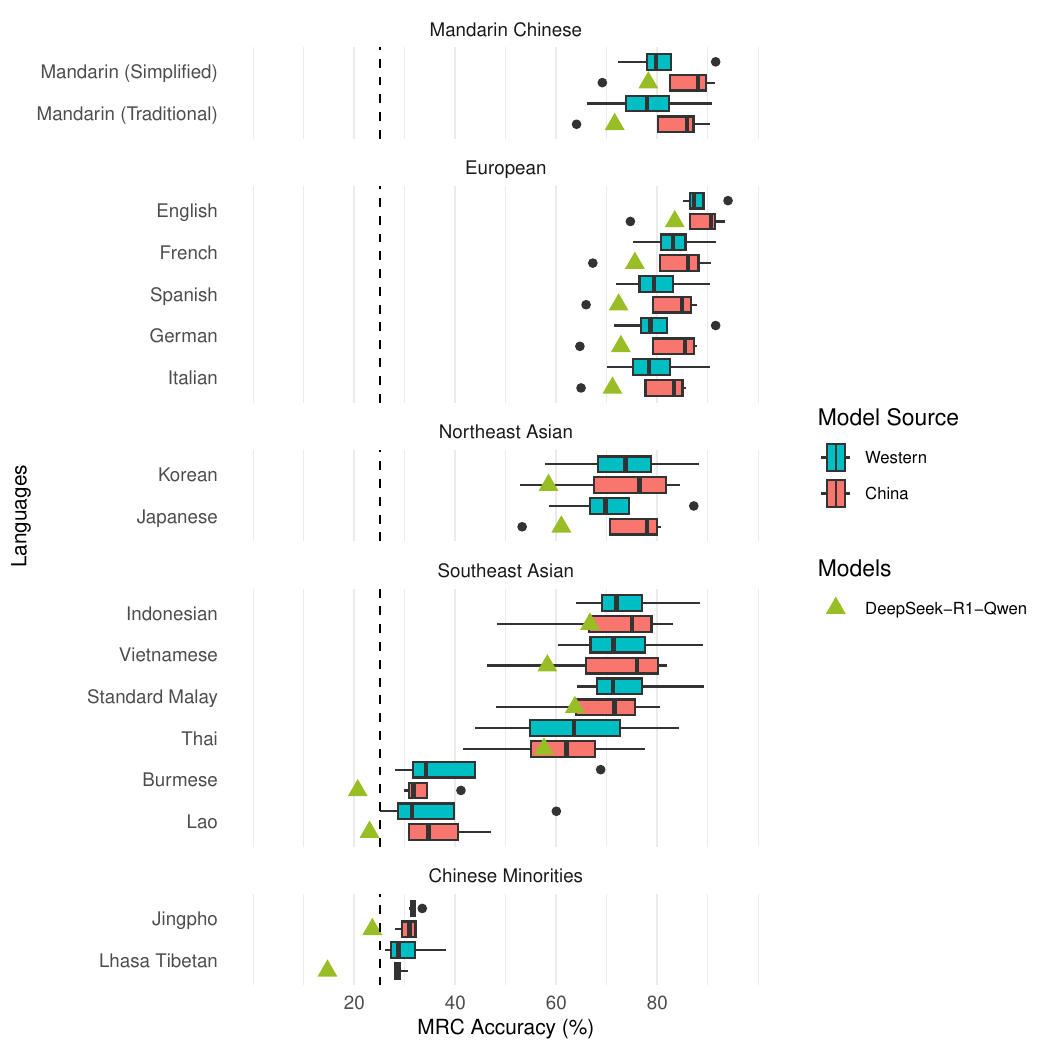}
    \caption{We presented the result of \texttt{DeepSeek-R1-Qwen} as green triangles to highlight its relations with other instruction-tuned models. All the models in the grouped bar results use the same chat template for comparability. However, we found that \texttt{DeepSeek-R1-Qwen} is particularly sensitive to chat template  (See Appendix Figure \ref{fig:chat-template-effect-MRC}). We therefore separately highlight its performance with the more effective chat template using a green triangle.}
    \label{fig:bele_whisker_full_chat_models_deepseekQwen}
\end{figure*}
\end{document}